\documentclass[review]{elsarticle}
\usepackage{lineno,hyperref}
\modulolinenumbers[5]

\makeatletter
\def\ps@pprintTitle{%
 \let\@oddhead\@empty
 \let\@evenhead\@empty
 \def\@oddfoot{}%
 \let\@evenfoot\@oddfoot}
\makeatother

\usepackage[utf8]{inputenc}  
\usepackage{listings}
\usepackage{color}
\usepackage{amssymb}  
\usepackage{amsmath}  
\usepackage{mathrsfs} 
\usepackage{bm}  
\usepackage{subcaption}  
\usepackage{setspace}

\biboptions{sort&compress} 

\usepackage{array}
\newcolumntype{L}[1]{>{\raggedright\let\newline\\\arraybackslash\hspace{0pt}}m{#1}}
\newcolumntype{C}[1]{>{\centering\let\newline\\\arraybackslash\hspace{0pt}}m{#1}}
\newcolumntype{R}[1]{>{\raggedleft\let\newline\\\arraybackslash\hspace{0pt}}m{#1}}
\newcolumntype{X}[1]{>{\raggedright\let\newline\\\arraybackslash\hspace{0pt}}p{#1}}
\newcolumntype{Y}[1]{>{\centering\let\newline\\\arraybackslash\hspace{0pt}}p{#1}}
\newcolumntype{Z}[1]{>{\raggedleft\let\newline\\\arraybackslash\hspace{0pt}}p{#1}}

\usepackage{siunitx}  
\makeatletter
\providecommand{\doi}[1]{%
	\begingroup
	\let\bibinfo\@secondoftwo
	\urlstyle{rm}%
	\href{http://dx.doi.org/#1}{%
		doi:\discretionary{}{}{}%
		\nolinkurl{#1}%
	}%
	\endgroup
}
\makeatother

\definecolor{cadmiumgreen}{rgb}{0.0, 0.42, 0.24}

\lstdefinestyle{floatcode}{float=tp,floatplacement=tbp
}

\newsavebox\CBox
\def\textBF#1{\sbox\CBox{#1}\resizebox{\wd\CBox}{\ht\CBox}{\textbf{#1}}}

\newcommand\descriptor{\mathcal{D}}
\newcommand\comparator{\mathcal{C}}
\newcommand\ranker{R}
\newcommand\rank{\tau}
\newcommand\rankof[1]{\rank_{#1}}
\newcommand\rankposition[2]{\rho_{#1}(#2)}

\newcommand\ranksof[1]{\mathcal{T}_{#1}}
\newcommand\scoresymbol{\varsigma}

\newcommand\scorein[3]{\varsigma_{#1}(#2,#3)}

\newcommand\FG{\mathcal{G}}
\newcommand\FGset{\mathbb{G}}
\newcommand\FGProjector{\mathcal{E}}
\newcommand\FV{\mathcal{V}}
\newcommand\FVset{\mathbb{V}}
\newcommand\real{\rm I\!R}

\journal{Journal of Pattern Recognition}

\begin{document}

\sloppy  

\begin{frontmatter}

\title{Fusion Vectors: Embedding Graph Fusions for Efficient Unsupervised Rank Aggregation}

\author[address1]{Icaro Cavalcante Dourado\corref{correspondingauthor}}
\ead{icaro.dourado@ic.unicamp.br}

\author[address1]{Ricardo da Silva Torres}
\ead{rtorres@ic.unicamp.br}

\cortext[correspondingauthor]{Corresponding author. Av. Albert Einstein, 1251, Campinas, SP, Brazil.}

\address[address1]{Institute of Computing, University of Campinas (UNICAMP), Campinas, Brazil}

\begin{abstract}

The increase in amount and complexity of digital content led to a wider demand of ad-hoc retrieval.
Complementary, heterogeneous data sources and retrieval models stimulated the proliferation of ingenious and effective rank aggregation functions.
Although recent rank aggregation functions are promising with respect to effectiveness, they usually overlook efficiency aspects.
We propose an innovative rank aggregation function that is unsupervised, intrinsically multimodal, and targeted for fast retrieval and top effectiveness performance.
We introduce embedding and indexing of graph-based rank-aggregation representation models, and applications for retrieval.
We propose embedding formulations for these representations.
We propose fusion vectors, a late-fusion representation based on ranks, from which a retrieval model is defined.
We propose fast retrieval based on fusion vectors, promoting an efficient rank aggregation system.
Our method presents top effectiveness performance among state of the art, while promoting multimodality and efficiency.
Consistent speedups are achieved against recent baselines.

\end{abstract}


\end{frontmatter}


\section{Introduction}

Huge volumes of complex data, comprising multiple kinds of modalities, have been created continuously. This scenario increases the demand of two research venues: (1) effective and efficient retrieval methods, and (2) the creation of sophisticated feature extraction algorithms.

Effective and efficient retrieval models should be employed to address existing users' information needs. One common solution relies on ad-hoc retrieval. Ad-hoc retrieval, also called content-based retrieval, allows documents, images, or multimodal objects to be adopted as queries in a search system. Ad-hoc retrieval has been exploited in several applications, such as service providers, digital libraries, and social media. Content-based image retrieval (CBIR)~\cite{torres:2006:CBIR,zhou:2017:CBIR} is an example of a common application field.

On the other side, feature extraction algorithms are important as they are the basis of subsequent generalization and learning models, commonly used in several domains, such as search and classification tasks.
Proposals of description approaches for images, texts, and multimedia data have advanced in the last decades, leading to more discriminative and effective models.
However, the choice of the most suitable technique often depends on the circumstances (e.g., application or dataset) in which they are used. In fact, an active research venue relies on exploiting their complementary view, by aggregation, aiming to improve the effectiveness of complex services, such as search, classification, or recommendation.

Existing aggregation methods are often categorized as early fusion or late fusion approaches. \textit{Early-fusion} methods emphasize the generation of composite descriptions for samples, whereas \textit{late-fusion} methods perform a combination of techniques focused on a target problem. Majority voting of classifiers and rank aggregation functions are examples of late-fusion methods.
Late-fusion methods are especially useful when the raw data from the objects are not available, and are potentially more effective than early-fusion methods because they are specifically designed or optimized for the problem being solved.

Rank aggregation functions allow retrieval models (or rankers) to be built on top of others.
They combine results from different rankers and promote more effective retrieval results, without dealing with raw data or low-level descriptors.
Besides, even heterogeneous models such as text-based or image-based can be gathered together.
Rank aggregation techniques are important in many applications, such as meta-search, document filtering, recommendation systems, and social choice.

Many rank aggregation functions have been proposed under varied approaches,
based on supervised learning~\cite{Mourao:2018:LowComplexityL2R,vargas:2015:L2Rgp},
rank position averaging~\cite{borda:1781:bordaCount,cormack2:2009:RRF,Montague:2002:Condorcet},
retrieval score combination~\cite{fox:1994:combination,pedronette:2013:rlsim},
Markov Chains~\cite{PaperSimilarItems_Sculley2007,RankAggregForWeb_WWW01}, or
graph of correlations~\cite{pedronette:2018:reciprocalGraph,zhang:2015:queryRankFusion}.
Among these works, graphs have been proved to be a powerful tool for modeling the relationships among data objects.
Notably, a novel graph-based rank aggregation approach has been recently proposed~\cite{Dourado:2019:FG}, which defined a rank-based representation model capable of encoding ranks relationships and which allows either retrieval or other tasks to be defined over that representation model~\cite{Dourado:2019:fusionGraphPrediction}.

Although some of these aggregation functions are promising with respect to effectiveness, many of them do not handle multimodality or have not been validated in such scenarios~\cite{Liang:2018:ManifoldRankAgg,pedronette:2018:reciprocalGraph,bai:2016:sparse,xie:2015:image}.
Yet, very few works have already investigated the proposal or representation models in the context of rank aggregation functions~\cite{Dourado:2019:FG,bai:2016:sparse}.
Besides, even recent proposals are not strictly bundled with efficiency~\cite{pedronette:2018:reciprocalGraph,Dourado:2019:FG}. Nevertheless, information retrieval typically has to deal with large datasets, thus demanding efficient retrieval.
On the other hand, a number of works from related research fields have been proposed regarding indexing structures, embedding formulations~\cite{cai:2018:graphEmbedding,Silva:2018:BoG,Dourado:2019:BoTG}, and approximate search~\cite{malkov:2018:HNSW}. We investigate the applicability of such initiatives in the context of rank aggregation, while targeting multimodality and represention models.

We propose an unsupervised rank aggregation function that provides both an effective and efficient retrieval system, also capable of dealing with heterogeneous rankers and multimodality.
Our solution adopts an underlying representation that encodes multiple ranks, with respect to a query, as a graph. An embedding approach is performed to project those graphs as vectors in a vector space model. Then, an indexing mechanism stores the resulting vectors from the collection, in order to provide time-efficient query resolution. The solution is overall unsupervised, so that no labeled data are required.

Such underlying vector-based representation of objects based on ranks even defines a new paradigm, in which not only could retrieval models be defined upon those models, but also they could act as higher-level representation structures for any task~\cite{Dourado:2019:fusionGraphPrediction}.

These are the contributions of the work:
\begin{enumerate}
\item The proposal of an innovative rank aggregation function, that it is unsupervised, intrinsically multimodal, and targeted for fast retrieval and top effectiveness performance;
\item The introduction of embedding approaches for graph-based rank-aggregation representation models;
\item An strategy for indexing and approximate retrieval based on rank-aggregation vector representations is presented.
\end{enumerate}

The remainder of this paper is organized as follows. Section~\ref{secRW} discusses related work and Section~\ref{secPrelimDefs} formally introduces preliminary definitions. Section~\ref{secMethod} presents the proposed method, while Section~\ref{secExpEval} describes conducted experiments. Finally, Section~\ref{secConc} concludes the paper and raises future research directions.

\section{Related Work}
\label{secRW}

Rank aggregation problem (RAP) targets an optimal rank that best represents a set of given base ranks.
RAP is NP-hard for more than three input ranks~\cite{RankAggregForWeb_WWW01}.
From a theoretical perspective, some heuristics have been proposed to find approximate solutions, based on, among other strategies, branch-and-bound search~\cite{Amodio:2016:accurateMedianRanking} and evolutionary algorithms~\cite{Aledo:2018:rankAggEvolution,Kaur:2017:rankAggMetasearchGenetProg,vargas:2015:L2Rgp,DAmbrosio:2017:EvolutionMedianRanking}.

In the context of information retrieval, RAP targets a permutation of retrieved objects obtained from different input ranks, hopefully more relevant to the user query than the base ranks.
Rank aggregation functions are then applied over a set of rankers to deliver better ranks in response to the users' needs.

Unsupervised rank aggregation functions work without relying on labeled training data. For that, they can be based on data discrimination or summarization strategies, such as rank position averaging~\cite{cormack2:2009:RRF,borda:1781:bordaCount,Montague:2002:Condorcet},
retrieval score combination~\cite{pedronette:2013:rlsim,fox:1994:combination}, correlation analysis~\cite{pedronette:2018:reciprocalGraph,zhang:2015:queryRankFusion}, or clustering~\cite{Liang:2018:ManifoldRankAgg}.

Existing graph-based methods are mostly targeted at modelling the whole collection of objects as a graph, from which the ranks can be derived~\cite{pedronette:2018:reciprocalGraph,zhang:2015:queryRankFusion,pedronette:2016:CorGraph,PaperSimilarItems_Sculley2007,pedronette:2016:RkGraph}.
Among these graph-based methods, a state-of-the-art unsupervised rank aggregation approach has been proposed recently, based on graph-based rank aggregation representations followed by similarity computations over those structures~\cite{Dourado:2019:FG}. This initiative -- referred to here as \textit{FG} -- was novel in establishing an aggregation representation, and overcame most previous baselines in retrieval effectiveness.
Its shortcoming, as in other previous works, is about the retrieval performance, as the query retrieval times are asymptotic linear to the collection size.
In this paper, we adopt \textit{FG} as our main baseline, for effectiveness and efficiency comparative analysis.

Parallel to those efforts, there are initiatives from fields such as database systems regarding indexing structures, embedding formulations~\cite{Silva:2018:BoG,cai:2018:graphEmbedding,Dourado:2019:BoTG}, and approximate search~\cite{malkov:2018:HNSW}.

Graph embedding approaches have been effective in multiple scenarios involving graph databases, because vector representations from graphs usually promote better scalability, and they also have more existing mining and search functions at disposal.
An embedding acts as a mapping function from a graph domain to a multidimensional vector space. \citet{Zhu:2014} proposed a map that roughly preserves distances between those domains, but to spaces of high dimensionality. This map can also be based on statistics of vertex attributes and edge attributes~\cite{Gibert:2012:graphEmbedding}, prototypes~\cite{bunke:2011:graphPrototypesSelection}, or graph kernels~\cite{Silva:2018:BoG}.
Among these initiatives, Bag of Graphs (BoG)~\cite{Silva:2018:BoG} was introduced as a general unsupervised framework for graph embedding that allows graphs to be represented as vectors based on common local graph patterns. Despite BoG is targeted for any graph scenario, it requires some functions to explicitly defined concerning the target scenario. BoG has been extended for some scenarios already, such as for text classification and retrieval~\cite{Dourado:2019:BoTG}.
\citet{He:2015:embeddingLearning} proposed a learning method for embedding representations of entities and relations previously modeled by graphs.

Complementary, indexing mechanisms have been extensively studied in the information retrieval literature, aiming at performing query retrieval efficiently by means of either exact or approximate nearest neighborhood search. These solutions usually adopt space partitioning~\cite{maneewongvatana:1999:kdtree}, hashing~\cite{gionis:1999:LSH}, or greedy search in neighborhood graphs~\cite{malkov:2018:HNSW}.

To our knowledge, this paper introduces one of the first initiatives that investigate those indexing and embedding approaches in the context of rank aggregation functions.
Our method promotes an efficiency aspect not previously covered by former rank aggregation initiatives, while also targeting effectiveness performance.

\section{Preliminary definitions}
\label{secPrelimDefs}

Here we formally establish preliminary definitions based on which we design our method. A few definitions are presented as defined by~\citet{Dourado:2019:FG}, such as {\em descriptor}, {\em comparator}, {\em ranker}, and {\em rank aggregation function}, while others are given for the first time.

A digital object, or {\em sample} $s$, can be represented by a {\em descriptor} $\descriptor$, as a vector, matrix, graph, or any other data structure: $\descriptor: s \mapsto \epsilon$. Each descriptor follows its own priors concerning the way it captures information from the raw data.

A {\em comparator} $\comparator$, where $\comparator: (\epsilon_i,\epsilon_j) \mapsto \scoresymbol \in \real^+$, is adopted to compare two samples $(s_i,s_j)$ in terms of their descriptions. Both underlying similarity or dissimilarity functions can be used, and we follow a procedure to convert dissimilarity into similarity scores, so that heterogeneous comparators can be used when applied together in a broader context.

Let $S$ be a \textit{response set} composed of \textit{response items} (samples), which are retrieved by an information retrieval system.

A \textit{ranker} $\ranker(\descriptor,\comparator)$, composed of a tuple of a descriptor $\descriptor$ and a comparator $\comparator$, or only $\ranker$ for short, establishes a ranking model such that $\ranker: q \mapsto \rank$.
It computes a \textit{rank} $\rank$ for the \textit{query} $q$, regarding $S$.
A rank is a permutation of $S_L \subseteq S$, where $L \ll n$ in general, such that $\rankof{q}$ provides the most similar -- or equivalently the least dissimilar -- response samples from $S$, to $q$, in order. $L$ is used as a cut-off parameter.
$\rankposition{\rankof{q}}{x}$ denotes the position of $x$ in $\rankof{q}$, starting by~$1$.

A ranker is also referred to as a retrieval model~\cite{baeza:1999:modernIR}.
In this sense, a {\em search} returns a rank $\rank$ in response of a query $q$.
In ad-hoc retrieval, $q$ follows the same definition of a sample, but refers to the input object in the context of a search.
An {\em exact search} -- for the query $q$, dataset $S$, and ranker $\ranker(\descriptor,\comparator)$ -- is asymptotically linear to $|S|$, and can be expressed by Equation~\ref{eq:exactSearch}, where \textit{argsort} is a function that returns the indices that would sort an array. In practice, only a top-$L$ of the response items most similar to $q$ are of interest.
{\em Approximate search}, conversely, works in sublinear time to $|S|$.
However, it has a trade-off between recall and complexity: it must be faster than the exact search, while retaining quality as much as possible.
It usually adopts {\em indexing structures} to reduce the search space~\cite{malkov:2018:HNSW}.
The main idea is that some loss is acceptable to make searches faster, specially for large datasets.
\begin{equation}\label{eq:exactSearch}
\rankof{q} = \underset { s_i \in S } { \operatorname { argsort } } \left( \comparator \left( \descriptor \left( q \right), \descriptor \left( s_i \right) \right) \right)
\end{equation}

Given $m$ rankers, $\{R_1,R_2,\ldots,R_m\}$, being used for retrieval over $S$, we can obtain for $q$ a rank set $\ranksof{q}= \{\rankof{1},\rankof{2},\ldots,\rankof{m}\}$, from which a \textit{rank aggregation function} $f: \ranksof{q} \mapsto \rankof{q,f}$ produces a combined rank, expected to be more effective than the individual ranks from $\ranksof{q}$.
We refer to {\em multi-ranked object} as an object represented by its multiple ranks. Different from early-fusion techniques, it is not represented by features from multiple feature extractors, but from ranks as if the object was a query over multiple rankers.

A {\em graph embedding function} $\FGProjector$ defines a $d$-dimensional vector space and projects graphs on it. A graph $\FG(V,E)$, where $V$ is the vertex set and $E$ is the edge set, is projected to that space as a vector $\FV$, so that $\FGProjector: \FG \mapsto \FV$, $d \ll |V|$ ideally, and $\FV \in \real^d$.
$\FGProjector$ is expected to preserve some graph properties and also proximity measures in the resultant space.

\section{Fast Rank Aggregation Retrieval}
\label{secMethod}

\begin{figure}[ht]
\includegraphics[width=\linewidth]{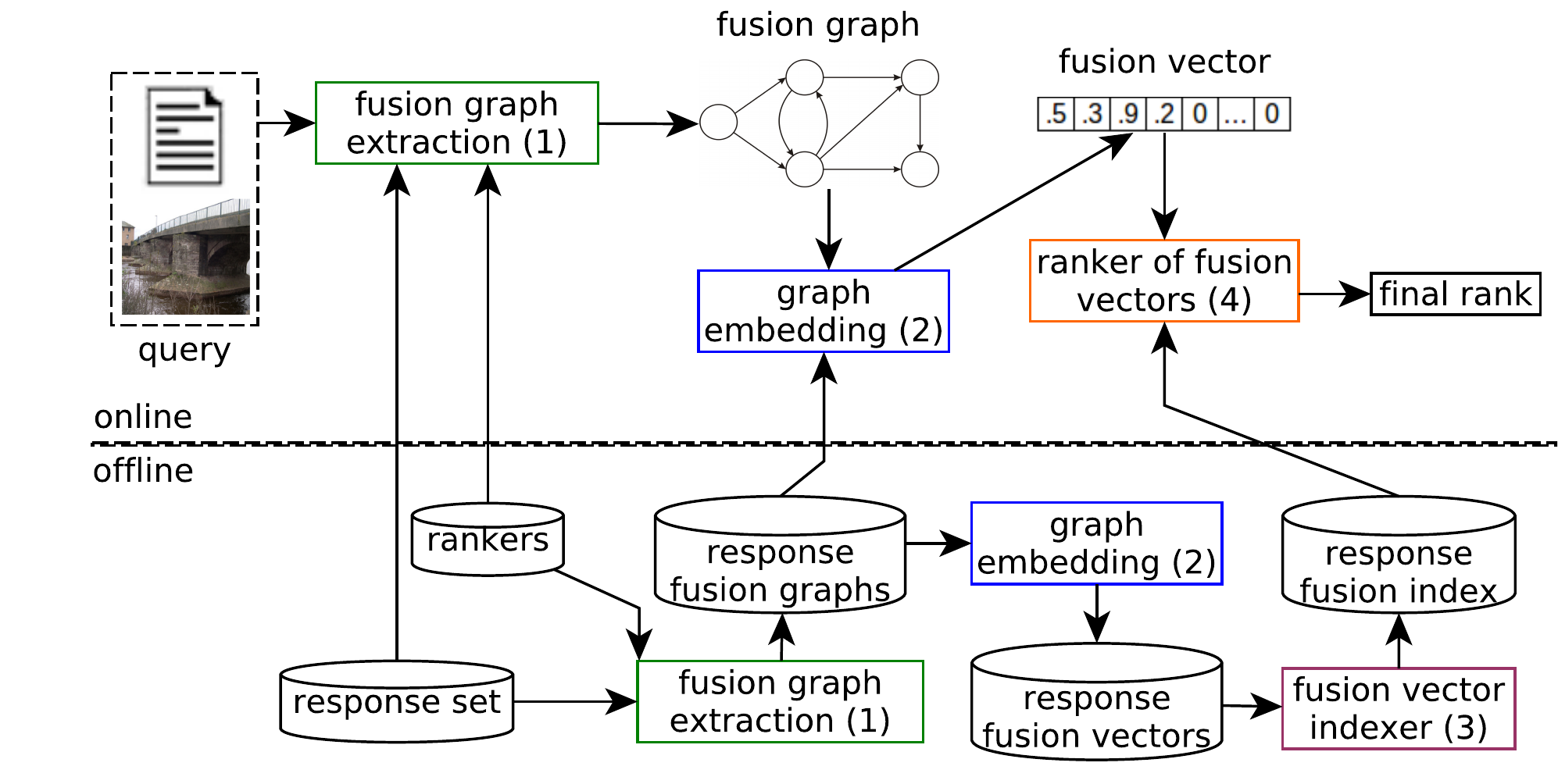}
\caption{Schematic view of the proposed method.}\label{fig:fusionVectorRetrievalFast}
\end{figure}

An overview of our rank aggregation proposal is shown in Figure~\ref{fig:fusionVectorRetrievalFast}, which highlights two stages -- offline and online -- and four numbered components.
The offline stage comprehends the modeling of the response set in terms of multiple rankers, and its indexing for further retrieval. This is performed only once. The online stage refers to the steps employed in a search session.
The four main generic components are briefly described here and detailed in the following sections.
The first two components are used in both stages.

The {\em fusion graph extraction} component (1) generates a fusion graph for a given query sample. A fusion graph $\FG$ consists of an aggregated representation of multiple ranks for a query, thus capturing and correlating information of its associated multiple ranks. This formulation is presented in Section~\ref{sec:fusionGraphExtraction}.

{\em Graph Embedding} (2) projects fusion graphs into a vector space model, producing a corresponding fusion vector $\FV$ per fusion graph. 
We propose and discuss some alternative embedding formulations in Section~\ref{sec:fusionGraphEmbedding}.

{\em Fusion vector indexer} (3) generates an index of fusion vectors, from which it is possible to retrieve response items for a given query, as long as both the query and the response items are previously represented as fusion vectors.
By means of the response fusion index, efficient searches of multi-ranked objects can be performed.
Although the ranks could be generated from the response fusion vectors directly, through brute-force search, the indexing step is important to promote sub-linear query processing time.

At the end, a {\em ranker of fusion vectors} (4) produces a rank of objects for a certain query object, according to the similarity of their respective fusion vectors.
These last two components, {\em fusion vector indexer} and {\em ranker of fusion vectors}, are detailed in Section~\ref{sec:fusionVectorIndexSearch}.

\subsection{Fusion Graph Extraction}
\label{sec:fusionGraphExtraction}

This component produces a fusion graph $\FG$ for a given query sample $q$ based on its ranks.
A fusion graph is a graph-based encoding of multiple ranks for $q$, that intrinsically encapsulates and correlates ranks.
We follow the fusion graph formulation from~\citet{Dourado:2019:FG}, who defined a procedure to extract a fusion graph $\FG$, for $q$, based on its ranks and ranks' inter-relationships. They also defined a retrieval model based on fusion graphs, hereby referred to as \textit{FG}. We adopt \textit{FG} as a baseline.

Let $\ranksof{q}$ be a set of $m$ ranks for the query $q$, with sizes up to a certain limit $L$, and obtained which respect to $m$ rankers, over a dataset $S$ of size $n$. Besides, consider that the ranks from every response item $s_i \in S$, regarding the $m$ rankers, are pre-computed in an offline stage.
Also, let $\scorein{\rankof{q}}{s_i}{s_j}$ be the similarity score between $s_i$ and $s_j$ with respect to the same descriptor $\descriptor$ and comparator $\comparator$ from the ranker $\ranker(\descriptor,\comparator)$ that produced $\rankof{q}$ for $q$.

The fusion graph extraction is a mapping function $\ranksof{q} \mapsto \FG$, and works in $O(m^2L^2)$.
$\FG$, for an object $q$, includes all response items from each rank $\rankof{q} \in \ranksof{q}$, as vertices.
Vertices are connected by taking into account the degree
of relationship between their corresponding response items, and the degree of their relationships to $q$.
The weight of a vertex $v_A$, expressed by $w({v_A})$, is given by Equation~\ref{eq:Wv}.
The weight of an edge $e_{A,B}$, expressed by $w({e_{A,B}})$, is given by Equation~\ref{eq:We}.

\begin{equation}
\label{eq:Wv}
w({v_A}) = \sum_{A \in \rankof{i} \wedge \rankof{i} \in \ranksof{q}} \scorein{\rankof{i}}{q}{A}
\end{equation}	

\begin{equation}
\label{eq:We}
w({e_{A,B}}) = \sum_{ A \in \rankof{i} \wedge \rankof{i} \in \ranksof{q}} \sum_{B \in \rankof{j} \wedge \rankof{j} \in \ranksof{A}} \left( \scorein{\rankof{j}}{A}{B} \div \rankposition{\rankof{i}}{A} \right)
\end{equation}

\subsection{Fusion Graph Embedding}
\label{sec:fusionGraphEmbedding}

Let $\FGset=\{\FG_1,\FG_2,\ldots,\FG_n\}$ be the fusion graph set related to the response set of a certain collection.
From $\FGset$, a fusion graph embedding function $\FGProjector$ defines a vector space in order to project a fusion graph $\FG(V,E)$ into that space as a fusion vector $\FV$, i.e., $\FGProjector: \FG \mapsto \FV$.

A fusion vector is a representation of multi-ranked objects, and allows efficient storage and search, as vectors are commonly much easier to manage than graphs. Dissimilarity scores between fusion vectors can be obtained by traditional vector comparators, such as Jaccard, cosine, or Euclidean functions. Based on fusion vectors, a retrieval system of multi-ranked objects can be further established.

$\FGProjector$ can be defined by unsupervised or supervised approaches. We focus on unsupervised approaches. We propose and evaluate three possible formulations, each one targeting different embedding categories~\cite{cai:2018:graphEmbedding}: vertex-based, hybrid, and kernel-based. Let $w_{\FG}(v)$ be the weight of the vertex $v$, if $v \in \FG$, otherwise 0.
Similarly, let $w_{\FG}(e)$ be the weight of the edge $e$, if $e \in \FG$, otherwise 0.
Also, let $d$ be the dimensionality of the vector space model defined by $\FGProjector$ and expressed by a \textit{d}-sized attribute set called \textit{vocabulary}, such that $\FV \in \real^d$, and $n = |\FGset|$.

\begin{figure}[ht]
\includegraphics[width=\linewidth]{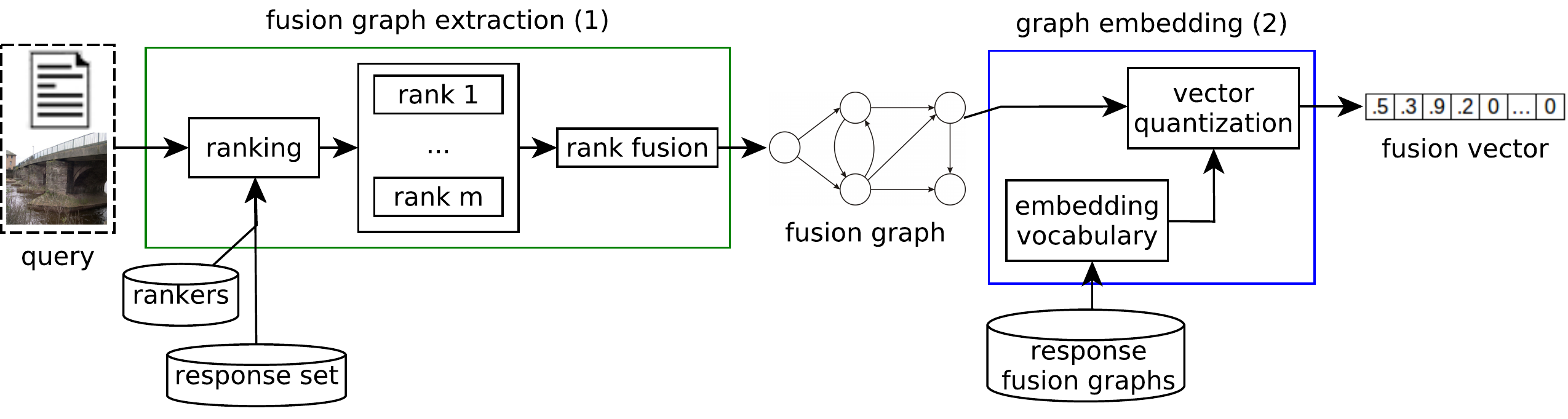}
\caption{Generation of a fusion vector.}\label{fig:fusionVectorFromSample}
\end{figure}

$\FGProjector$ first defines (or learns) its own embedding vocabulary, in order to apply it to generate the fusion vectors in a step called \textit{vector quantization}.
Figure~\ref{fig:fusionVectorFromSample} illustrates the generation of a fusion vector, detailing the integration between the components \textit{fusion graph extraction} (1) and \textit{graph embedding} (2).

\subsubsection{Vertex-based Embedding}

$\FGProjector_V$ is the first and simplest of our proposed embedding formulations, which derives $\FV$ from the vertices of $\FG$.
For $\FGProjector_V$, there is one vector attribute relative to each response object, therefore $d=n$, and a fusion vector is defined as
\begin{equation} \label{eq:fvVector}
\FV=(u_1,\ldots,u_i,\ldots,u_d),
\end{equation}
where $1 \le i \le d$, $u_i=w_{\FG}(v_i)$.

Despite that vector space increases linearly to the collection size, the resulting fusion vectors are mainly sparse, i.e., composed of few non-zero entries, which allows this embedding formulation to be efficient for storage and for dissimilarity comparisons.

\subsubsection{Hybrid Embedding}

$\FGProjector_{H}$ is an embedding formulation that, different from $\FGProjector_V$, derives the fusion vector from both the vertices and edges of $\FG$, therefore called a \textit{hybrid} embedding. In $\FGProjector_{H}$, each response object contributes to one attribute in the vector space. Besides, each possible edge linking two distinct vertices, $e(v_i,v_j)$, contributes to an additional vector attribute, but we handle inverted pairs -- $e(v_i,v_j)$ and $e(v_j,v_i)$ -- to refer to the same attribute, as if the edges were undirected. Hence, the vector space has dimensionality
\begin{equation} \label{eq:fvHybridDimensions}
d = n + (\frac{n^2}{2} - n) = \frac{n^2}{2}.
\end{equation}
The fusion vector is defined as
\begin{equation} \label{eq:fvHybrid}
\FV=(u_1,\ldots,u_i,\ldots,u_n,x_1,\ldots,x_k,\ldots,x_m),
\end{equation}
where $1 \le i \le n$, $1 \le k \le m$, $m=\frac{n^2}{2}-n$,
$u_i=w_{\FG}(v_i)$, $i < j$, and $x_k=w_{\FG}(e_{v_i,v_j})+w_{\FG}(e_{v_j,v_i})$.

$\FGProjector_{H}$ has the benefit over $\FGProjector_V$ in incorporating proximity information, at a cost of leading to a representation with more dimensions.

\subsubsection{Kernel-based Embedding}

In a kernel-based embedding, a graph is represented as a vector containing the frequencies of elementary substructures that are decomposed from that graph~\cite{cai:2018:graphEmbedding}.
In this context, a graph kernel defines an atomic substructure, such as a subgraph of fixed size (graphlet), a subtree pattern, or a random walk.

$\FGProjector_K$ is the third proposed embedding formulation, which extends Bag of Graphs (BoG)~\cite{Silva:2018:BoG} -- a kernel-based embedding framework -- to the rank aggregation domain. To the best of our knowledge, this is the first work that extends BoG to this scenario.

BoG is a general framework for graph embedding, but requires some functions to be explicitly defined according to the target scenario. The vector space is defined by a vocabulary named \textit{codebook}, which is a set of attributes called \textit{codewords}. Codewords are common local graph patterns, based on subgraphs either selected as centroids of a subgraph clustering procedure or by random selection.
The main idea is that graphs can be projected to vectors computed as histograms of subgraphs.
We extend BoG in order to promote $\FGProjector_K$, as indicated in Figure~\ref{fig:bog}, using the following definitions, where $\mathscr{G}=\{g_1, g_2, \dots, g_{|\mathscr{G}|}\}$ means a set of subgraphs from one or more fusion graphs:

\begin{figure}
\centering
\includegraphics[width=0.8\textwidth]{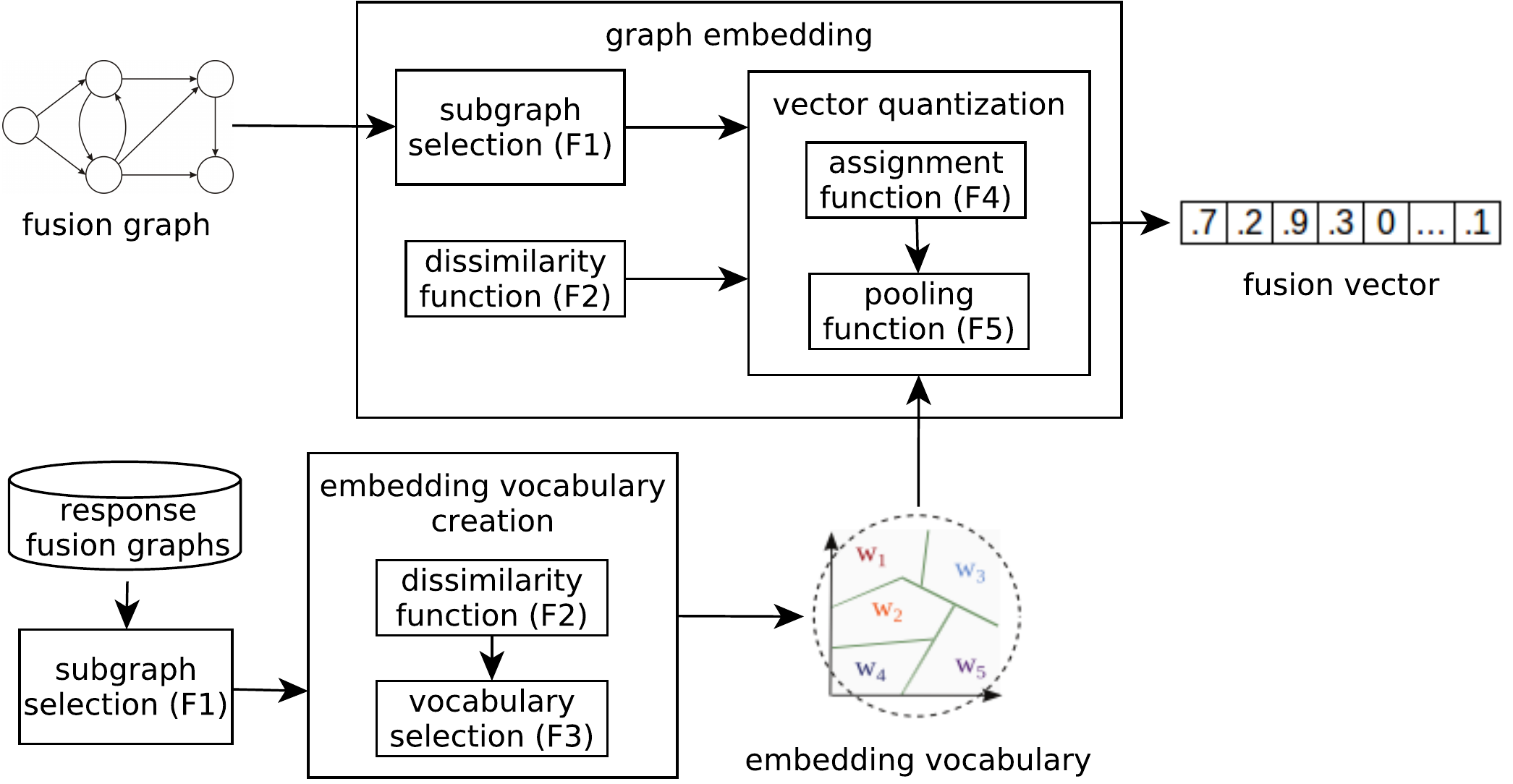}
\caption{Kernel-based embedding of fusion graphs as an extension of the BoG framework.}
\label{fig:bog}
\end{figure}

\begin{enumerate}              
\item Subgraph selection (\textit{F1}): function $\mathcal{P}(G) \to \{0,1\}$ that determines if a subgraph of $\FG$ is of interest, i.e., if it satisfies a property $P$.
Graph of Interest (GoI) is a pattern of $\FG$, so that a set of valid subgraphs of $\FG$ can be extracted. We adopt the following subgraph definition: for every vertex $v \in \FG$, we derive one subgraph containing: (1) $v$; (2) all direct incident vertices starting from $v$; and (3) the edges linking them. We preserve both vertex weights and edge weights into the subgraph.

\item Graph dissimilarity function (\textit{F2}): it is implemented trough the use of a GoI. F2 is a function $\delta : (g_a,g_b) \mapsto \scoresymbol \in \real^+$ that computes the similarity between two graphs.
We adopt MCS~\cite{Bunke:1998:MCS}, which computes the dissimilarity score based on maximum common subgraphs, and can be efficiently implemented linearly on the number of vertices~\cite{Dourado:2019:BoTG}.

\item Vocabulary selection (\textit{F3}): Let $\mathscr{G}$ be a set of subgraphs from multiple fusion graphs, F3 computes a partition on $\mathscr{G}$. The resulting sets are clusters. A codebook $\mathfrak{C}=\{w_1, w_2, \dots, w_{|\mathfrak{C}|}\}$ is a set of codewords representing each group defined by a clustering.
    
\item Assignment function (\textit{F4}): Given $\mathfrak{C}$, and $\mathscr{G}$ obtained for a certain input fusion graph $\FG$, F4 defines an activation value for each pair $(g_i,w_j)$, where $g_i \in \mathscr{G}$ and $w_j \in \mathfrak{C}$.
We adopt Soft Assignment, expressed in Equation~\ref{eq:soft_assignment}~\cite{Dourado:2019:BoTG}, which employs a Gaussian to establish smooth scores for an input subgraph $g_i$ to each graph attribute $w_j$~\cite{Silva:2018:BoG}, where ${K(x) = \frac{exp(-\frac{x^2}{2\sigma^2})}{\sigma\sqrt{2\pi}}}$, and $\sigma$ allows the smoothness control.

\item Pooling function (\textit{F5}): Let $f_{assign}$ be an assignment function.
A coding is $C=\{ c_1, c_2, \dots, c_{|\mathscr{G}|}\},$ where $c_i$ is a vector that $c_{i}[j] = f_{assign}(g_i,w_j)$, where $g_i \in \mathscr{G}$ and
$w_j \in \mathfrak{C}$, $1 \leq i \leq |\mathscr{G}|$, $1 \leq j \leq |\mathfrak{C}|$. Given $C$, $F5:C \mapsto \real^d$ is a function that summarizes all word assignments, defined in a coding $C$, into a numerical vector.
We adopt Average Pooling (Equation~\ref{eq:avg_pooling}~\cite{Dourado:2019:BoTG}), which weights the $j$-th vector attribute, $c[j]$, as the percentage of associations of the input sample subgraphs to the $j$-th graph attribute, $w_j$.

\end{enumerate}

\begin{equation}
\label{eq:soft_assignment}
c_i[j] = \frac{K(\delta(g_i, w_j))}{sum_{k=1}^{d} K(\delta(g_i, w_k))}
\end{equation}

\begin{equation}
    \label{eq:avg_pooling}
    c[j] =  \frac{ \sum_{i=1}^{|\mathcal{G}|} c_i[j] }{ |\mathcal{G}| }
\end{equation}

$\FGProjector_K$ has the potential to produce more discriminative and compact embeddings than $\FGProjector_V$ and $\FGProjector_{H}$, but requires additional computation, domain specialization, and adjustment of hyperparameters.

\subsection{Index and Search of Fusion Vectors}
\label{sec:fusionVectorIndexSearch}

Fusion vectors not only do act as a representation of multi-ranked objects, but also allows their retrieval, thus promoting intrinsic rank aggregation.
This section shows how fusion vectors can be used to promote rank-aggregation by means of an efficient retrieval through indexing and approximate search.

Let $\FVset=\{\FV_1,\FV_2,\ldots,\FV_n\}$ be the fusion vector set related to the response set of a certain collection. A {\em fusion vector indexer} creates an index of $\FVset$, in order to allow efficient searches of fusion vectors in response of a query fusion vector, as previously shown in Figure~\ref{fig:fusionVectorRetrievalFast}.

We index fusion vectors by extending the Hierarchical Navigable Small World (HNSW)~\cite{malkov:2018:HNSW}, a state-of-the-art archetype that enables approximate K-nearest neighbor searches while retaining high recall rates. HNSW builds a graph of neighbors, a connected graph of controllable hierarchy and restricted number of edges that allows efficient searches by a greedy heuristic.
We essentially extend the HNSW reference implementation\footnote{\url{https://github.com/nmslib/hnswlib}} in order to promote indexing and search of our fusion vectors -- which are sparse vectors -- along with support for the cosine dissimilarity, so that we could make it work for high dimensional and sparse data at a small memory consumption. These contributions have already been incorporated into its open-source code.

\section{Experimental Evaluation}
\label{secExpEval}

We present, in this section, the proposed evaluation protocol and the experimental results achieved by our rank aggregation function, in contrast to results from individual rankers and related works.
We evaluate it regarding effectiveness and efficiency aspects.

\subsection{Datasets and Features}

We evaluate our proposal comprising searching scenarios over public datasets of diverse purposes, in order to validate it for general applicability.
The datasets are listed in Table~\ref{tab:datasets}, along with the individual rankers adopted to generate ranks for aggregation.
These datasets are commonly adopted in evaluations of recent related works~\cite{Dourado:2019:FG,pedronette:2018:reciprocalGraph,zhang:2015:queryRankFusion,bai:2016:sparse,Valem:2018:RankCorkNNSets}, due to exemplifying various real retrieval tasks, such as by content, texture, shape, color, textual, and multimodal.
The rankers were selected according to their purposes and the objective involved in each dataset.
We evaluate the effectiveness of individual rankers, to serve as a first baseline, as well as their rank aggregation with methods proposed in the literature.

\textbf{UKBench}~\cite{nister:2006:UKBench} is a dataset of \num{10200} images, consisting of \num{2550} scenes/objects captured four times each. These captures vary in terms of illumination, viewpoint, and distance. The objects/scenes correspond to the categories, being four samples per class. The effectiveness assessment in UKBench relies on the N-S Score evaluation metric, which varies from zero to four, and measures the mean number of relevant images among the first four images retrieved.
We adopt seven rankers, some based on color and texture properties, and also deep-learning-based rankers. In Table~\ref{tab:datasets}, CNN-Caffe~\cite{Jia:2014:CNNCaffe} stands for the $4096$-dimensional output from the $7$th layer of a
Convolution Neural Network (CNN) obtained with the Caffe framework, plus the Euclidean distance as comparator.

\textbf{Ohsumed}~\cite{hersh:1994:ohsumed} is a bibliographic collection from the National Library of Medicine, consisting of \num{34389} abstracts of cardiovascular diseases distributed across $23$ categories. We use the subset of \num{18302} uni-labeled documents, varying from $56$ to $2876$ documents per category.
We preprocess the documents with stop word removal and Porter's stemming. The rankers adopted are based on Bag-of-words (BoW), 2grams, graph-based models, and language models based on \textit{word embeddings}. For BoW and 2grams, rankers are composed with either cosine and Jaccard comparators.

\textbf{Brodatz}~\cite{brodatz:1966} is a dataset of \num{1776} images (texture blocks), being $16$ samples for each of the $111$ classes (texture types). We adopt three texture rankers.

\textbf{MPEG-7}~\cite{latecki:2000:shape} is a shape dataset of \num{1400} images, distributed in $20$ images per $70$ categories. We adopt six shape rankers.

\textbf{Soccer}~\cite{van:2006:soccer} is an image dataset of $280$ images, distributed in $40$ images per $7$ categories.
We adopt three color-based rankers.

University of Washington (\textbf{UW})~\cite{Deselaers:2008:UW} is a hybrid dataset, composed of \num{1109} scene pictures annotated by textual keywords, and distributed across $20$ classes, varying from $22$ to $255$ pictures per class. The keywords per picture vary from $1$ to $22$.
We adopt twelve rankers, comprising six textual rankers, three visual color rankers, and three visual texture rankers.

\begin{table}[ht]
\centering
\caption{Datasets and rankers used in the experimental evaluation.}
\label{tab:datasets}
\resizebox{.99\columnwidth}{!}{\begin{tabular}{lX{15cm}X{2.5cm}}
\hline
\textbf{Dataset} & \textbf{Rankers} & \textbf{Ranker Types}\\
\hline
UKBench & ACC~\cite{huang:1997:ACC}, VOC~\cite{wang:2011:VOC}, SCD~\cite{manjunath:2001:SCD}, JCD~\cite{zagoris:2010:JCD}, CNN-Caffe~\cite{Jia:2014:CNNCaffe}, FCTH-SPy~\cite{chatzichristofis:2008:FCTH,lux:2011:irLIRe}, CEDD-SPy~\cite{chatzichristofis:2008:CEDD,lux:2011:irLIRe} & Color, Texture, BoVW, CNN \\
Ohsumed & BoW-cosine, BoW-Jaccard, 2grams-cosine, 2grams-Jaccard, GNF-MCS~\cite{schenker2007:graphtext,Bunke:1998:MCS}, GNF-WGU~\cite{schenker2007:graphtext,Wallis:2001:WGU}, WMD~\cite{kusner:2015:WMD} & Textual \\
Brodatz & LBP~\cite{ojala:2002:LBP}, CCOM~\cite{kovalev:1998:CCOM}, LAS~\cite{tao:2000:LAS} & Texture \\
MPEG-7 & SS~\cite{torres:2007:SS}, BAS~\cite{arica:2003:BAS}, IDSC~\cite{ling:2007:IDSC}, CFD~\cite{pedronette:2010:CFD}, ASC~\cite{ling:2010:ASC}, AIR~\cite{gopalan:2010:AIR} & Shape \\
Soccer & GCH~\cite{swain:1991:GCH}, ACC~\cite{huang:1997:ACC}, BIC~\cite{stehling:2002:BIC} & Color \\
UW & GCH~\cite{swain:1991:GCH}, BIC~\cite{stehling:2002:BIC}, JAC~\cite{williams:2007:JAC}, HTD~\cite{wu:1999:HTD}, QCCH~\cite{huang:2007:QCCH}, LAS~\cite{tao:2000:LAS}, COSINE~\cite{baeza:1999:modernIR}, JACCARD~\cite{lewis2006textsimilarityan}, TF-IDF~\cite{baeza:1999:modernIR}, DICE~\cite{lewis2006textsimilarityan}, OKAPI~\cite{robertson:1995:okapi}, BOW~\cite{carrillo2009representing} & Color, Texture, Textual \\
\hline
\end{tabular}}
\end{table}

\subsection{Experimental Protocol}

The first evaluation intends to analyze the effectiveness of our rank aggregation function in retrieval scenarios, compared to individual rankers and other aggregation functions. Second, we analyze the efficiency and trade-offs of our embedding approaches and indexed formulations.

We perform a protocol for object retrieval, also referred to as ad-hoc retrieval, which is commonly employed to evaluate rank aggregation methods.
Given the datasets used,
we treat each sample $s$ as query $q$ at a time, whose result candidates belong to $S$. A retrieved item is relevant to $q$ if they belong to the same class, since we are validating in labeled collections, i.e., relevance scores are either $1$ for relevant or $0$ for irrelevant. In this case, the query set size corresponds to the dataset size.
In general, separate query and response sets can be used, as well as graded relevance.

We measure the retrieval effectiveness by the normalized discounted cumulative gain at cutoff 10 (NDCG@10) for all datasets except UKBench, for which we use the N-S Score, the standard measure in this dataset.

Three possible combinations of rankers are evaluated per dataset, as in~\cite{Dourado:2019:FG}: all rankers; the two most effective rankers; and the pair that maximizes a trade-off measure between high effectiveness and low correlation. We adopt multiple evaluation scenarios per dataset to provide a comprehensive analysis and to allow comparisons of different ranker selection strategies for rank aggregation.

We run and evaluate, using the same experimental procedure, the following baselines:
FG~\cite{Dourado:2019:FG},
QueryRankFusion~\cite{zhang:2015:queryRankFusion},
RecKNNGraphCCs~\cite{pedronette:2018:reciprocalGraph},
RkGraph~\cite{pedronette:2016:RkGraph},
CorGraph~\cite{pedronette:2016:CorGraph},
MRA~\cite{fagin:2003:MRA},
and RRF~\cite{cormack2:2009:RRF}.
Other related works could also be included, but we focus on the most recent and competitive unsupervised approaches. \citet{Dourado:2019:FG} empirically confirmed a number of related works that are no longer competitive to state-of-the-art methods, so that they can be now suppressed in future benchmarks.
That comprehends methods such as CombSUM, CombMIN, CombMAX, CombMED, CombANZ, and CombMNZ~\cite{fox:1994:combination}, BordaCount~\cite{borda:1781:bordaCount}, Condorcet~\cite{Montague:2002:Condorcet}, Kemeny~\cite{kemeny:1959:Kemeny}, and RLSim~\cite{pedronette:2013:rlsim}.
For UKBench, as it imposes its own evaluation protocol and evaluation metric, we can also compare our results against state-of-the-art methods on this dataset, regardless of their various underlying approaches. In this sense, our method can be compared to early-fusion approaches, supervised rank-fusion approaches, etc.

We compare the \textit{winning number}~\cite{Tax:2018:benchmarkLtK} of each rank aggregation function. This allows us to compare multiple methods concerning several datasets, fusion configurations and baselines.
The winning number of a method $m$, $W_m$, is a global performance indicator for a performance measure $P$, expressed by Equation~\ref{eq:winningNumber}, where
$D$ is the set of datasets,
$C_d$ is the set of our $3$ pre-defined configurations for dataset $d$ with respect to the rankers to fuse,
$M$ is set of rank aggregation methods,
$P_m(d,c)$ is the performance of $m \in M$ on $d \in D$ and configuration $c \in C_d$,
and $\bm{1}_{P_m(d,c) > P_k(d,c)}$ is given by Equation~\ref{eq:indicatorFunction}.

\begin{equation} \label{eq:winningNumber}
W_m = \sum_{d \in D} \sum_{c \in C_d} \sum_{i \in M} 1_{P_m(d,c) > P_i(d,c)}
\end{equation}

\begin{equation}\label{eq:indicatorFunction}
\bm{1}_{P_m(d,c) > P_i(d,c)} =
\begin{cases} 
1 & \text{if $P_m(d,c) > P_i(d,c)$}, \\
0 & \text{otherwise.}
\end{cases}
\end{equation}


The embedding approaches and the indexing scheme, defined before, hold different trade-offs to the retrieval tasks. For this reason, besides effectiveness, we also analyze the efficiency of our method comprising its alternative formulations, and also compare to \textit{FG}, the main baseline.
We measure the mean time spent per query, for each combination of rankers in each dataset.
The elapsed time for query retrieval refers to the sum of the times spent for fusion graph extraction, fusion vector extraction (graph embedding), and object retrieval, as represented by the online stage in Figure~\ref{fig:fusionVectorRetrievalFast}.
The mean time of 5 independent measurements is reported.
For the baseline \textit{FG}, the same steps but the fusion vector extraction (absent) is taken into consideration.

In the presentation and discussion about results, we assume the acronyms indicated in Table~\ref{tab:methodAcronyms}.
There are variants regarding the embedding approach and dissimilarity function, as well as the use of the indexing mechanism.

\begin{table}[ht]
\centering
\caption{Acronyms of the method variants.}
\label{tab:methodAcronyms}
\begin{tabular}{ll}\hline
Acronym & Meaning \\\hline


\textit{FV-V} & Fusion vector generated by $\FGProjector_V$ \\
\textit{FV-H} & Fusion vector generated by $\FGProjector_{H}$ \\
\textit{FV-K} & Fusion vector generated by $\FGProjector_K$ \\
\textit{FV-V-FAST} & The indexed counterpart of \textit{FV-V} \\
\textit{FV-H-FAST} & The indexed counterpart of \textit{FV-H} \\
\textit{FV-K-FAST} & The indexed counterpart of \textit{FV-K} \\

\hline
\end{tabular}
\end{table}

\subsection{Ranker Effectiveness}

The effectiveness of the rankers are shown in Tables~\ref{tab:rankersResultsBrodatz},
\ref{tab:rankersResultsUW},
\ref{tab:rankersResultsMPEG7},
\ref{tab:rankersResultsOhsumed},
\ref{tab:rankersResultsUKBench}, and
\ref{tab:rankersResultsSoccer} for the datasets Brodatz, UW, MPEG-7, Ohsumed, UKBench, and Soccer, respectively.
These results serve as an initial baseline for the rank aggregation functions, so that the aggregation functions are expected to overcome them.

We can observe large variability in rankers' results.
Rankers' relative performance also vary depending on the dataset, thus providing complementary views.
JACCARD, for instance, was superior to COSINE in UW dataset, but was worse in Ohsumed.

\begin{table*}[ht]
\caption{Effectiveness of individual rankers on the datasets.}
\begin{subtable}[t]{0.32\textwidth}
\centering
\caption{Brodatz}
\label{tab:rankersResultsBrodatz}
\resizebox{3.4cm}{!}{%
\begin{tabular}{lr}\hline
Ranker & NDCG@10  \\\hline
LAS        & \textBF{0.850533} \\
CCOM       & 0.726186 \\
LBP        & 0.652759 \\\hline
\end{tabular}}
\vspace{2mm}
\caption{UW dataset}
\label{tab:rankersResultsUW}
\resizebox{3.8cm}{!}{%
\begin{tabular}{lr}\hline
Ranker & NDCG@10  \\\hline
JAC & \textBF{0.810729} \\
BIC & 0.746454 \\
DICE & 0.722831 \\
BOW & 0.720781 \\
OKAPI & 0.716035 \\
JACCARD & 0.701651 \\
TF-IDF & 0.658880\\
GCH & 0.630315 \\
COSINE & 0.554767 \\
LAS & 0.514314 \\
HTD & 0.495002 \\
QCCH & 0.414249 \\\hline
\end{tabular}}
\end{subtable}
\begin{subtable}[t]{0.32\textwidth}
\centering
\caption{MPEG-7}
\label{tab:rankersResultsMPEG7}
\resizebox{3.5cm}{!}{%
\begin{tabular}{lr}\hline
Ranker & NDCG@10  \\\hline
ASC  & \textBF{0.941585} \\
AIR  & 0.939424 \\
CFD  & 0.930685 \\
IDSC & 0.922828 \\
BAS  & 0.866098 \\
SS   & 0.611481 \\\hline
\end{tabular}}
\vspace{2mm}
\caption{Ohsumed}
\label{tab:rankersResultsOhsumed}
\resizebox{3.8cm}{!}{%
\begin{tabular}{lr}\hline
Ranker & NDCG@10  \\\hline
BoW-cosine & \textBF{0.669701} \\
2grams-cosine & 0.664120 \\
GNF-WGU & 0.662668 \\
GNF-MCS & 0.655420 \\
2grams-Jaccard & 0.651320 \\
BoW-Jaccard & 0.645711 \\
WMD & 0.427361 \\\hline
\end{tabular}}
\end{subtable}
\begin{subtable}[t]{0.32\textwidth}
\centering
\caption{UKBench}
\label{tab:rankersResultsUKBench}
\resizebox{3.6cm}{!}{%
\begin{tabular}{lr}\hline
Ranker & N-S Score \\\hline
VOC & \textBF{3.54} \\
ACC & 3.37 \\
CNN-Caffe & 3.31 \\
SCD & 3.15 \\
JCD & 2.79 \\
FCTH-SPy & 2.73 \\
CEDD-SPy & 2.61 \\
\hline
\end{tabular}}
\vspace{2mm}
\caption{Soccer}
\label{tab:rankersResultsSoccer}
\resizebox{3.4cm}{!}{%
\begin{tabular}{lr}\hline
Ranker & NDCG@10  \\\hline
BIC & \textBF{0.614818} \\
ACC & 0.592699 \\
GCH & 0.536412 \\\hline
\end{tabular}}
\end{subtable}
\end{table*}

\subsection{Rank Aggregation Results}

The effectiveness of our method, for different embedding approaches and indexed formulations, along with the related works, are shown in
Tables~\ref{tab:aggResultsUKBench},
\ref{tab:aggResultsOhsumed},
\ref{tab:aggResultsBrodatz},
\ref{tab:aggResultsMPEG7},
\ref{tab:aggResultsSoccer},
and \ref{tab:aggResultsUW}, respectively for UKBench, Ohsumed, Brodatz, MPEG-7, Soccer, and UW. Three ranker selections are evaluated per dataset.

\begin{table}[ht]
\centering
\caption{Results for rank aggregation on UKBench.}
\label{tab:aggResultsUKBench}
\resizebox{.7\columnwidth}{!}{\begin{tabular}{lZ{4.5cm}Z{0.9cm}Z{1.9cm}}\hline
Method & \multicolumn{3}{c}{N-S Score} \\\hline
& VOC + ACC + CNN-Caffe + SCD + JCD + FCTH-SPy + CEDD-SPy & VOC + ACC & VOC + ACC + CNN-Caffe \\\hline
\textbf{FV}-V-FAST    & 3.74 & 3.86 & 3.92 \\
\textbf{FV}-H-FAST    & 3.74 & 3.86 & 3.92 \\
\textbf{FV}-V        & 3.74 & 3.86 & 3.92 \\
\textbf{FV}-H        & 3.74 & 3.86 & 3.92 \\
FG~\cite{Dourado:2019:FG} & 3.69 & 3.83 & 3.90 \\
RecKNNGraphCCs~\cite{pedronette:2018:reciprocalGraph} & 3.67 & 3.81 & 3.87 \\
QueryRankFusion~\cite{zhang:2015:queryRankFusion} & 3.60 & 3.78 & 3.86 \\
\textbf{FV}-K-FAST & 3.60 & 3.72 & 3.81 \\
\textbf{FV}-K     & 3.60 & 3.72 & 3.81 \\
MRA~\cite{fagin:2003:MRA} & 3.52 & 3.50 & 3.77 \\
RRF~\cite{cormack2:2009:RRF} & 3.52 & 3.60 & 3.76 \\
RkGraph~\cite{pedronette:2016:RkGraph} & 3.03 & 3.50 & 3.54 \\
CorGraph~\cite{pedronette:2016:CorGraph} & 2.44 & 2.91 & 2.77 \\
\hline
\end{tabular}}
\end{table}

\begin{table}[ht]
\centering
\caption{Results for rank aggregation on Ohsumed.}
\label{tab:aggResultsOhsumed}
\resizebox{.75\columnwidth}{!}{
\begin{tabular}{lZ{5.5cm}Z{2.3cm}Z{2cm}}\hline
Method & \multicolumn{3}{c}{NDCG@10} \\\hline
 & BoW-cosine + BoW-Jaccard + 2grams-cosine + 2grams-Jaccard + GNF-MCS + GNF-WGU + WMD & BoW-cosine + 2grams-cosine & BoW-cosine + WMD \\\hline
FG~\cite{Dourado:2019:FG} & 0.683835 & 0.683472 & 0.676760 \\
RecKNNGraphCCs~\cite{pedronette:2018:reciprocalGraph} & 0.676234 & 0.679728 & 0.667750 \\

\textbf{FV}-H-FAST & 0.676325 & 0.678596 & 0.664745 \\
\textbf{FV}-H & 0.676310 & 0.678419 & 0.665320 \\
\textbf{FV}-V-FAST & 0.666759 & 0.672709 & 0.624808 \\
\textbf{FV}-V & 0.666511 & 0.672674 & 0.627062 \\

QueryRankFusion~\cite{zhang:2015:queryRankFusion} & 0.651279 & 0.671258 & 0.669704 \\
MRA~\cite{fagin:2003:MRA} & 0.666045 & 0.670357 & 0.582049 \\
RRF~\cite{cormack2:2009:RRF} & 0.665793 & 0.671294 & 0.571016 \\
\textbf{FV}-K & 0.596524 & 0.665829 & 0.502955 \\
\textbf{FV}-K-FAST & 0.545134 & 0.666344 & 0.502555 \\
CorGraph~\cite{pedronette:2016:CorGraph} & 0.487177 & 0.497431 & 0.456434 \\
RkGraph~\cite{pedronette:2016:RkGraph} & 0.289045 & 0.688443 & 0.288436 \\
\hline
\end{tabular}}
\end{table}

\begin{table}[ht]
\centering
\caption{Results for rank aggregation on Brodatz.}
\label{tab:aggResultsBrodatz}
\resizebox{.7\columnwidth}{!}{\begin{tabular}{lrrr}\hline
Method & \multicolumn{3}{c}{NDCG@10} \\\hline
 & LAS+CCOM+LBP & LAS+CCOM & LAS+LBP \\\hline
RecKNNGraphCCs~\cite{pedronette:2018:reciprocalGraph} & 0.877882 & 0.882903 & 0.839717 \\
\textbf{FV}-K-FAST & 0.883388 & 0.880668 & 0.839430 \\
\textbf{FV}-K & 0.883388 & 0.880666 & 0.839425 \\
FG~\cite{Dourado:2019:FG} & 0.878995 & 0.872084 & 0.835624 \\
\textbf{FV}-H-FAST & 0.867736 & 0.858995 & 0.825835 \\
\textbf{FV}-H & 0.867688 & 0.858960 & 0.825785 \\
\textbf{FV}-V-FAST & 0.863513 & 0.854987 & 0.821171 \\
\textbf{FV}-V & 0.863513 & 0.854909 & 0.821135 \\
RkGraph~\cite{pedronette:2016:RkGraph} & 0.812659 & 0.861250 & 0.788682 \\
QueryRankFusion~\cite{zhang:2015:queryRankFusion} & 0.850263 & 0.850438 & 0.808562 \\
RRF~\cite{cormack2:2009:RRF} & 0.818656 & 0.817139 & 0.788840 \\
MRA~\cite{fagin:2003:MRA} & 0.822778 & 0.813396 & 0.788883 \\
CorGraph~\cite{pedronette:2016:CorGraph} & 0.749420 & 0.895623 & 0.719204 \\
\hline
\end{tabular}}
\end{table}

\begin{table}[ht]
\centering
\caption{Results for rank aggregation on MPEG-7.}
\label{tab:aggResultsMPEG7}
\resizebox{.7\columnwidth}{!}{\begin{tabular}{lZ{3.3cm}Z{1.7cm}Z{1.5cm}}\hline
Method & \multicolumn{3}{c}{NDCG@10} \\\hline
 & AIR + CFD + ASC + IDSC + BAS + SS & ASC + AIR & AIR + CFD \\\hline
\textbf{FV}-K-FAST & 0.998322 & 0.997401 & 0.998182 \\
\textbf{FV}-K     & 0.998322 & 0.997401 & 0.998182 \\
RecKNNGraphCCs~\cite{pedronette:2018:reciprocalGraph} & 0.998052 & 0.995160 & 0.997267 \\
FG~\cite{Dourado:2019:FG} & 0.997658 & 0.994729 & 0.995886 \\
\textbf{FV}-H-FAST    & 0.996635 & 0.991278 & 0.991684 \\
\textbf{FV}-H        & 0.996635 & 0.991278 & 0.991684 \\
\textbf{FV}-V        & 0.996272 & 0.988776 & 0.987672 \\
\textbf{FV}-V-FAST    & 0.996272 & 0.988776 & 0.987671 \\
RkGraph~\cite{pedronette:2016:RkGraph} & 0.826119 & 0.999350 & 0.992078 \\
CorGraph~\cite{pedronette:2016:CorGraph} & 0.992456 & 0.962951 & 0.961460 \\
RRF~\cite{cormack2:2009:RRF} & 0.980638 & 0.957684 & 0.954499 \\
MRA~\cite{fagin:2003:MRA} & 0.980086 & 0.950442 & 0.946144 \\
QueryRankFusion~\cite{zhang:2015:queryRankFusion} & 0.940976 & 0.941762 & 0.941271 \\
\hline
\end{tabular}}
\end{table}

\begin{table}[ht]
\centering
\caption{Results for rank aggregation on Soccer.}
\label{tab:aggResultsSoccer}
\resizebox{.7\columnwidth}{!}{\begin{tabular}{lrrr}\hline
Method & \multicolumn{3}{c}{NDCG@10} \\\hline
 & BIC+ACC+GCH & BIC+ACC & BIC+GCH \\\hline
RkGraph~\cite{pedronette:2016:RkGraph} & 0.653623 & 0.656422 & 0.628563 \\
\textbf{FV}-K-FAST & 0.642172 & 0.656412 & 0.628528 \\
\textbf{FV}-K     & 0.641453 & 0.656173 & 0.628474 \\
FG~\cite{Dourado:2019:FG} & 0.651828 & 0.655332 & 0.622217 \\
\textbf{FV}-H-FAST    & 0.651308 & 0.650073 & 0.621168 \\
\textbf{FV}-H        & 0.650997 & 0.649573 & 0.620213 \\
\textbf{FV}-V-FAST    & 0.647671 & 0.648442 & 0.619718 \\
\textbf{FV}-V        & 0.647149 & 0.648151 & 0.619468 \\
CorGraph~\cite{pedronette:2016:CorGraph} & 0.645004 & 0.643505 & 0.623627 \\
RecKNNGraphCCs~\cite{pedronette:2018:reciprocalGraph} & 0.637537 & 0.640729 & 0.618704 \\
QueryRankFusion~\cite{zhang:2015:queryRankFusion} & 0.613732 & 0.613659 & 0.598862 \\
RRF~\cite{cormack2:2009:RRF} & 0.604119 & 0.613005 & 0.590819 \\
MRA~\cite{fagin:2003:MRA} & 0.605971 & 0.611017 & 0.588399 \\
\hline
\end{tabular}}
\end{table}

\begin{table}[ht]
\centering
\caption{Results for rank aggregation on UW.}
\label{tab:aggResultsUW}
\resizebox{.9\columnwidth}{!}{\begin{tabular}{lZ{9cm}Z{1.5cm}Z{1.5cm}}\hline
Method & \multicolumn{3}{c}{NDCG@10} \\\hline
 & JAC + BIC + DICE + BOW + OKAPI + JACCARD + TF-IDF + GCH + COSINE + LAS + HTD + QCCH & JAC + BIC & JAC + OKAPI \\\hline
CorGraph~\cite{pedronette:2016:CorGraph} & 0.896341 & 0.842665 & 0.933452 \\
\textbf{FV}-K-FAST & 0.888936 & 0.857554 & 0.899520 \\
\textbf{FV}-K     & 0.888770 & 0.857554 & 0.899256 \\
FG~\cite{Dourado:2019:FG} & 0.873607 & 0.854473 & 0.882776 \\
RecKNNGraphCCs~\cite{pedronette:2018:reciprocalGraph} & 0.869448 & 0.843423 & 0.882035 \\
\textbf{FV}-V-FAST    & 0.862406 & 0.841015 & 0.877076 \\
\textbf{FV}-H-FAST    & 0.862289 & 0.844613 & 0.874147 \\
\textbf{FV}-H        & 0.862108 & 0.844539 & 0.874078 \\
\textbf{FV}-V        & 0.862229 & 0.840928 & 0.877016 \\
RkGraph~\cite{pedronette:2016:RkGraph} & 0.746804 & 0.841127 & 0.866544 \\
MRA~\cite{fagin:2003:MRA} & 0.815983 & 0.797292 & 0.786995 \\
RRF~\cite{cormack2:2009:RRF} & 0.815779 & 0.798502 & 0.795143 \\
QueryRankFusion~\cite{zhang:2015:queryRankFusion} & 0.747281 & 0.792681 & 0.807250 \\
\hline
\end{tabular}}
\end{table}

\textit{FV} overcame \textit{FG}, the strongest and main baseline, in all $6$ datasets, but Ohsumed. Nevertheless, in Ohsumed, \textit{FV} surpassed all other baselines.



\textit{FV-K} was the most effective rank aggregation function, compared to \textit{FV-V} and \textit{FV-H}, in $4$ out of $6$ datasets. As expected, the kernel-based embedding performed better than the other two approaches, but at a higher computational cost.
Interestingly, \textit{FV-V} was better than \textit{FV-H} in $4$ out of $6$ datasets, which may be due to the large increase in terms of the vector dimensionality.

In UKBench, Brodatz, MPEG-7, Soccer, and UW, the indexed versions of all the three embeddings in all three aggregation scenarios promoted nearly the same effectiveness results than their non-indexed versions.
In Ohsumed, the indexed versions of FV-V and FV-H had nearly equivalent effectiveness to their non-indexed versions, and only for FV-K the indexing actually decreased the results (by 4\%) but only in one of three aggregation scenarios.
Regarding the effectiveness of the indexed formulations for FV, we can conclude that the indexing effectively contributes to our solution in terms of efficiency, while retaining the quality results in almost every case.

\begin{figure}
\centering
\begin{subfigure}{.5\textwidth}
\centering
\includegraphics[width=0.95\linewidth]{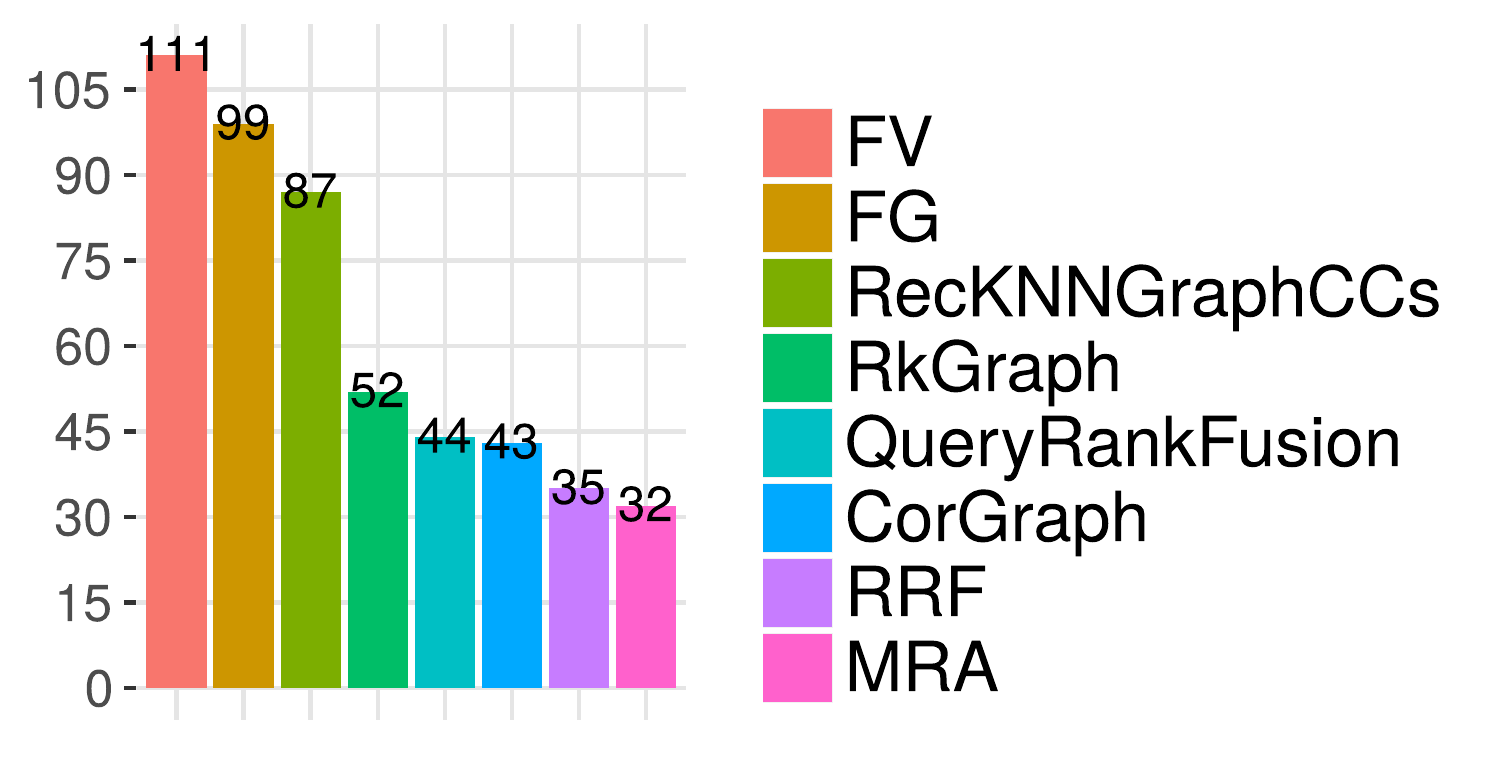}
\caption{Best instance per method.}
\label{fig:winningNumbersSimpler}
\end{subfigure}%
\begin{subfigure}{.5\textwidth}
\centering
\includegraphics[width=0.95\linewidth]{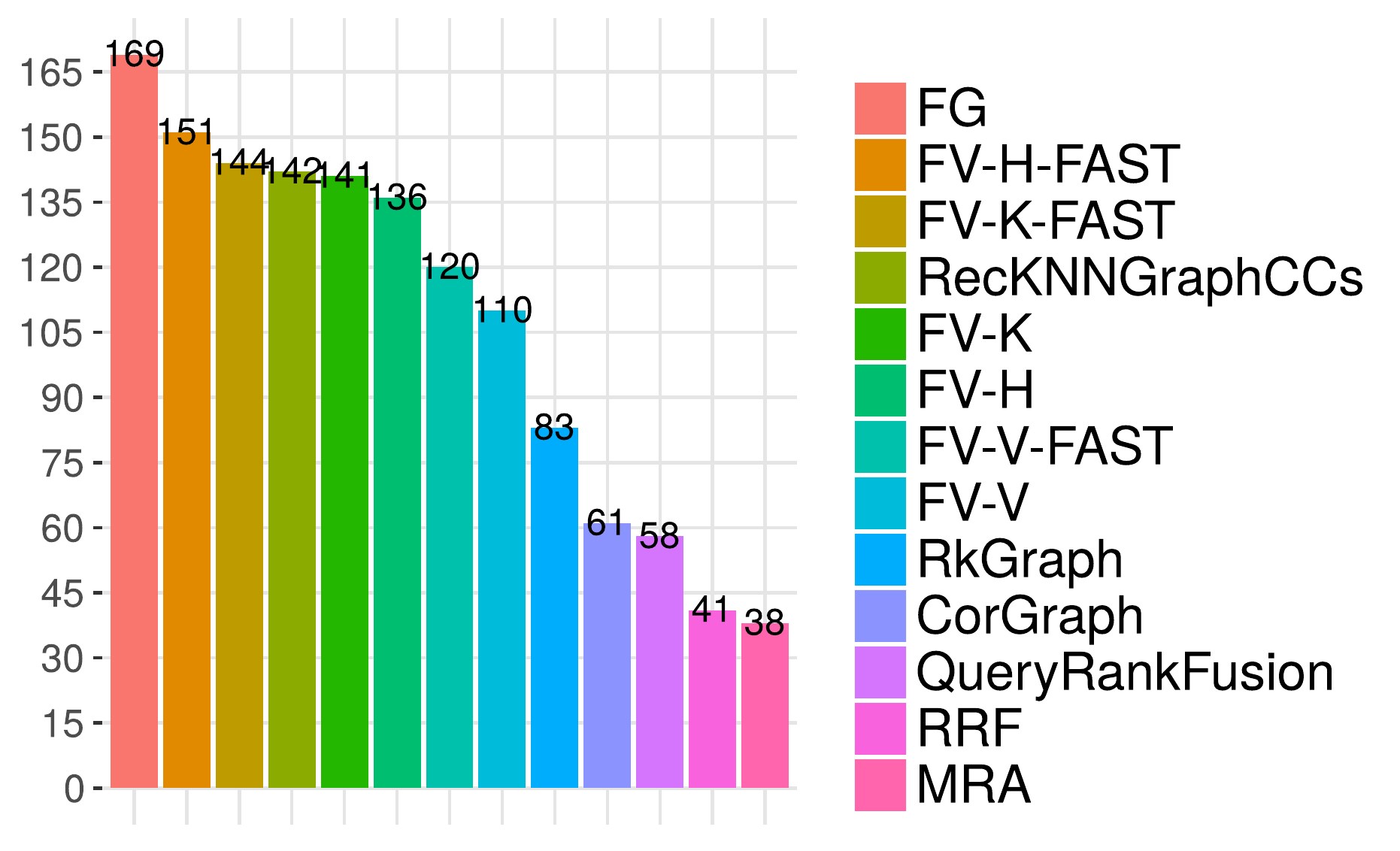}
\caption{All methods.}
\label{fig:winningNumbers}
\end{subfigure}
\caption{Winning numbers achieved per rank aggregation function.}
\label{fig:winningNumbersBoth}
\end{figure}

By comparing the aggregation functions globally in terms of winning numbers, \textit{FV} overcame all baselines, if we take the best result of its variants per aggregation per dataset, as shown in Figure~\ref{fig:winningNumbersSimpler}.
When we compare each \textit{FV} variant separately, along with the baselines, the best \textit{FV} approach is competitive and overcome all baselines but \textit{FG}, in terms of effectiveness (Figure~\ref{fig:winningNumbers}).
This is due to the unsupervised nature of the problem, which does not involve hyperparameter adjustment. The \textit{FV} approaches are competitive between each other, varying in which performs best per dataset. This issue, however, can be solved by additional pre-validation steps, if desired, in order to pick the most promising embedding. Besides, the efficiency benefits of our method can be of critical importance depending on the task. We investigate this trade-off in the following section.

Given that retrieval and fusion models, when validated in UKBench, follow its own evaluation procedure, we can present our results for it in contrast to the state of the art, besides the baselines we already reported in Table~\ref{tab:aggResultsUKBench}. These works involve both late-fusion~\cite{Valem:2018:RankCorkNNSets,xie:2015:image,bai:2016:sparse} and early-fusion~\cite{zheng:2016:multiScaleContextual,Yang:2015:Diffusion,wang:2012:metricfusion} approaches, and also even supervised approaches~\cite{zheng:2016:multiScaleContextual}.
Table~\ref{tab:artOfTheArtUKBench} reports the results, where those marked with * were obtained as the rank aggregation of the rankers ACC + VOC + CNN-Caffe.
QueryRankFusion is presented twice, one regarding their own reported result in~\cite{zhang:2015:queryRankFusion}, and another considering the same input rankers as ours.
Our method achieved the best performance, over other fusion approaches. Even compared to supervised approaches, which uses data that we do not rely on, our method shows high effectiveness.
Taking the effectiveness of the strongest isolated ranker as an initial baseline, which was VOC for UKBench as shown in Table~\ref{tab:rankersResultsUKBench} with a N-S score of $3.54$, two of our related works -- RkGraph and CorGraph -- did not surpass that baseline. In contrast, our method surpass results from isolated rankers in all evaluated scenarios.

\begin{table}
\centering
\caption{State-of-the-art results on UKBench. Methods marked with * were evaluated using ACC + VOC + CNN-Caffe.}
\label{tab:artOfTheArtUKBench}
\begin{tabular}{ll}\hline
Method & N-S Score \\
\hline
\textBF{FV*} & \textBF{3.92} \\
\citet{Valem:2018:RankCorkNNSets} & 3.92 \\
FG*~\cite{Dourado:2019:FG} & 3.90 \\
\citet{xie:2015:image} & 3.89 \\
\citet{zheng:2016:multiScaleContextual} & 3.88 \\
RecKNNGraphCCs*~\cite{pedronette:2018:reciprocalGraph} & 3.87 \\
\citet{bai:2016:sparse} & 3.86 \\
QueryRankFusion*~\cite{zhang:2015:queryRankFusion} & 3.86 \\
\citet{Yang:2015:Diffusion} & 3.86 \\
QueryRankFusion~\cite{zhang:2015:queryRankFusion} & 3.83 \\
MRA*~\cite{fagin:2003:MRA} & 3.77 \\
RRF*~\cite{cormack2:2009:RRF} & 3.76 \\
\citet{wang:2012:metricfusion} & 3.68 \\
RkGraph*~\cite{pedronette:2016:RkGraph} & 3.54 \\
CorGraph*~\cite{pedronette:2016:CorGraph} & 2.91 \\
\hline
\end{tabular}
\end{table}

\subsection{Efficiency Analysis}

\begin{figure}[t!]
\centering
\begin{subfigure}{.33\textwidth}
\centering
\includegraphics[width=.97\linewidth]{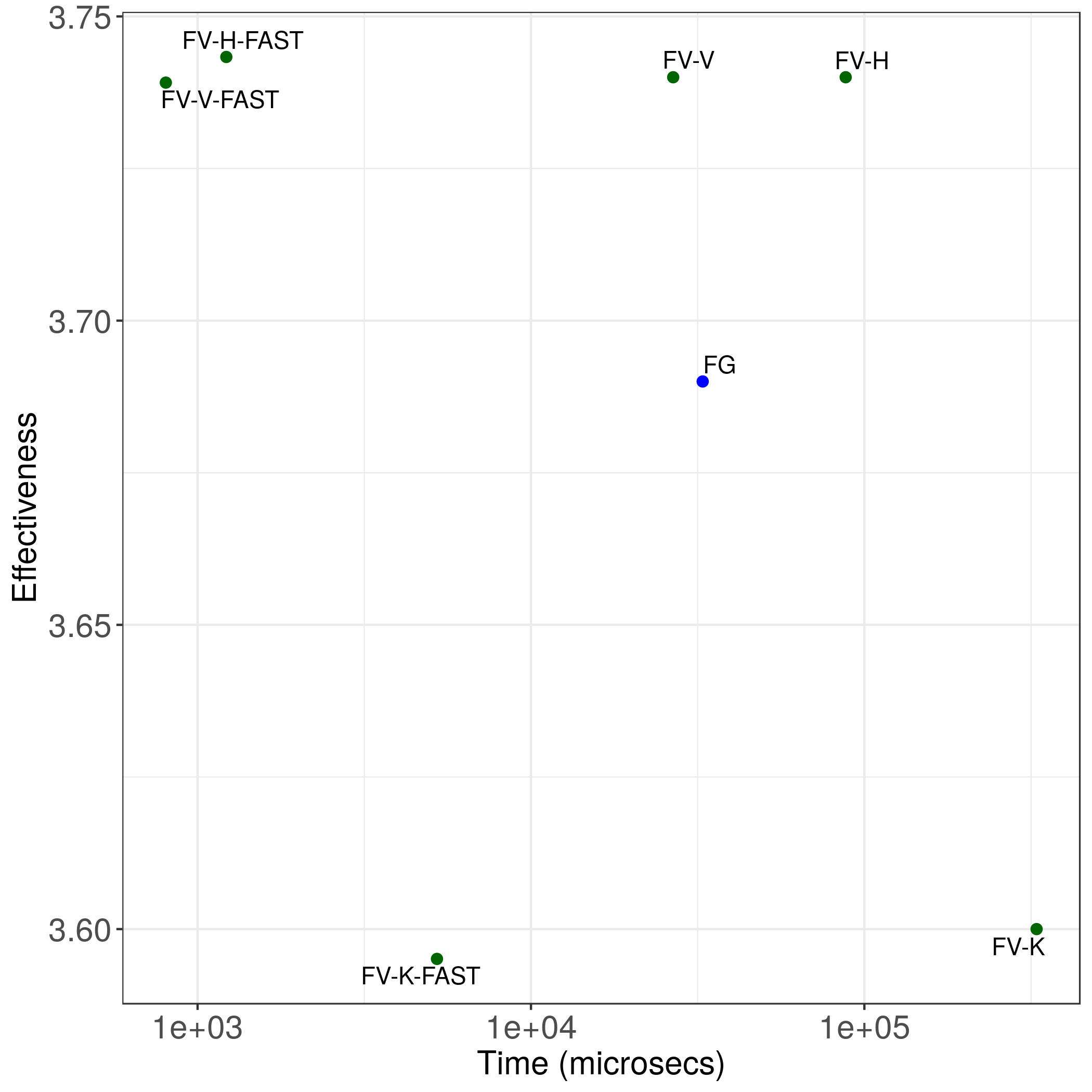}
\end{subfigure}
\begin{subfigure}{.33\textwidth}
\centering
\includegraphics[width=.97\linewidth]{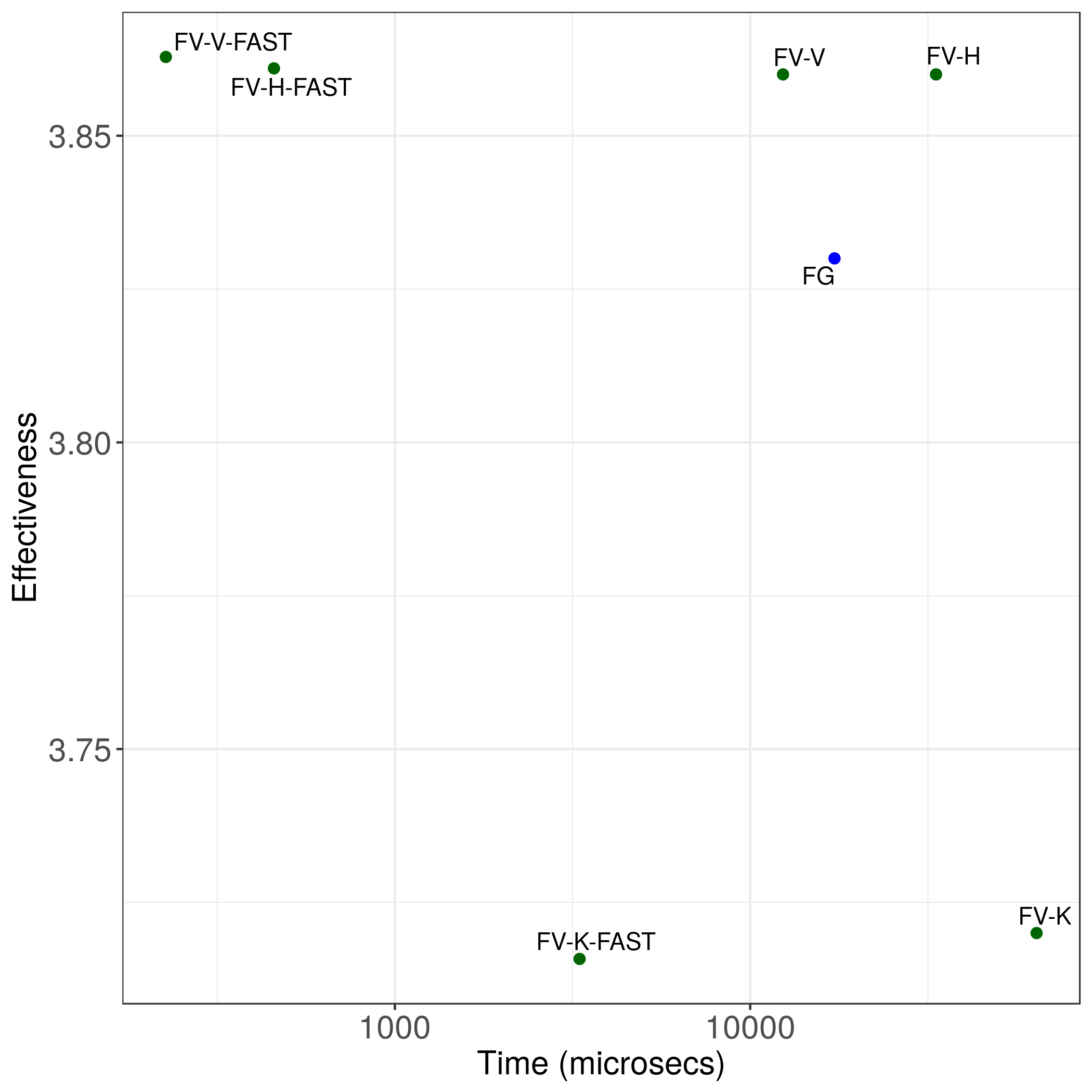}
\end{subfigure}
\begin{subfigure}{.33\textwidth}
\centering
\includegraphics[width=.97\linewidth]{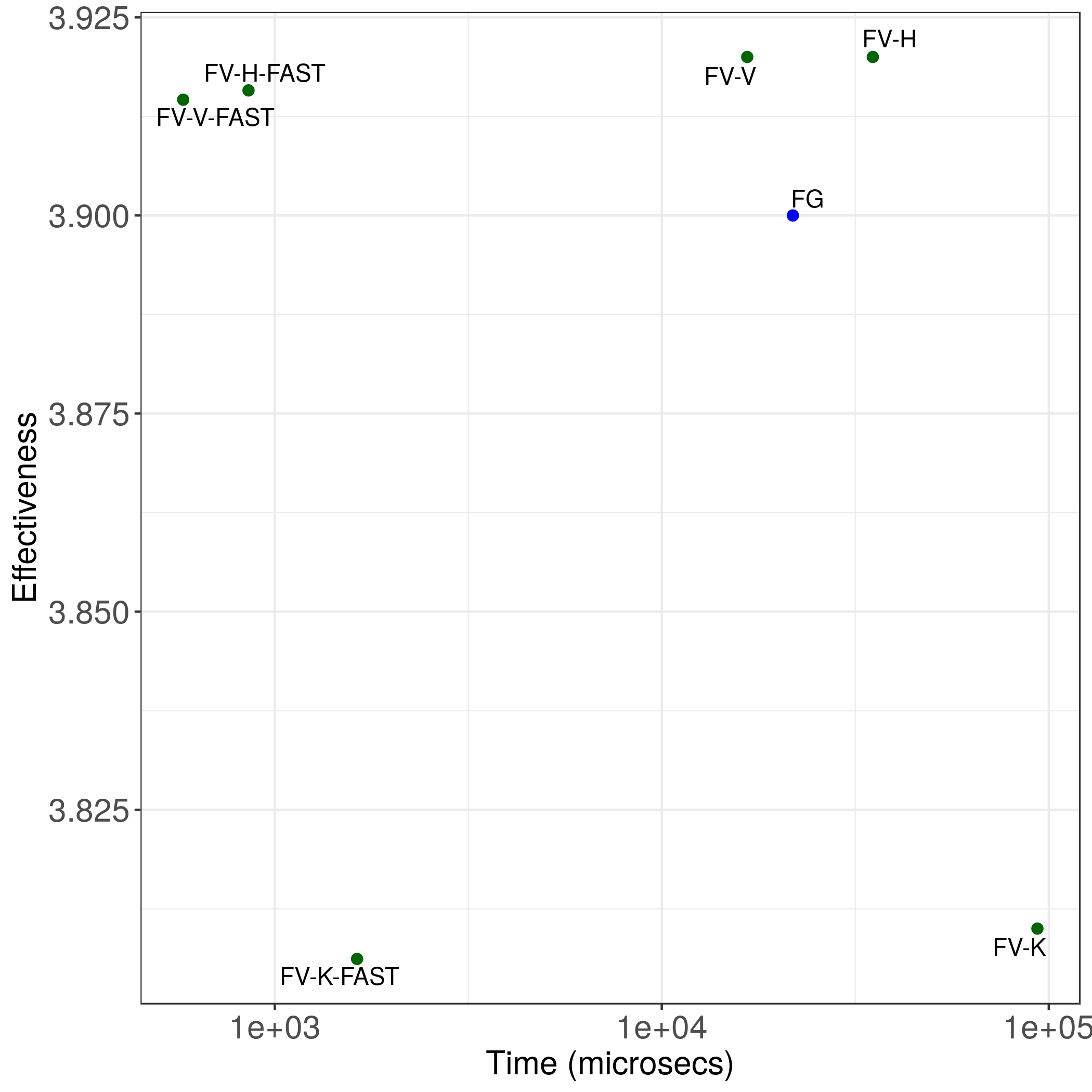}
\end{subfigure}
\caption{Effectiveness and efficiency trade-offs for FV and its embedding and indexed versions, in UKBench, for VOC + ACC + CNN-Caffe + SCD + JCD + FCTH-SPy + CEDD-SPy, VOC + ACC, and VOC + ACC + CNN-Caffe, respectively.}
\label{fig:tradeoff-ukbench}
\end{figure}

\begin{figure}[t!]
\centering
\begin{subfigure}{.33\textwidth}
\centering
\includegraphics[width=.97\linewidth]{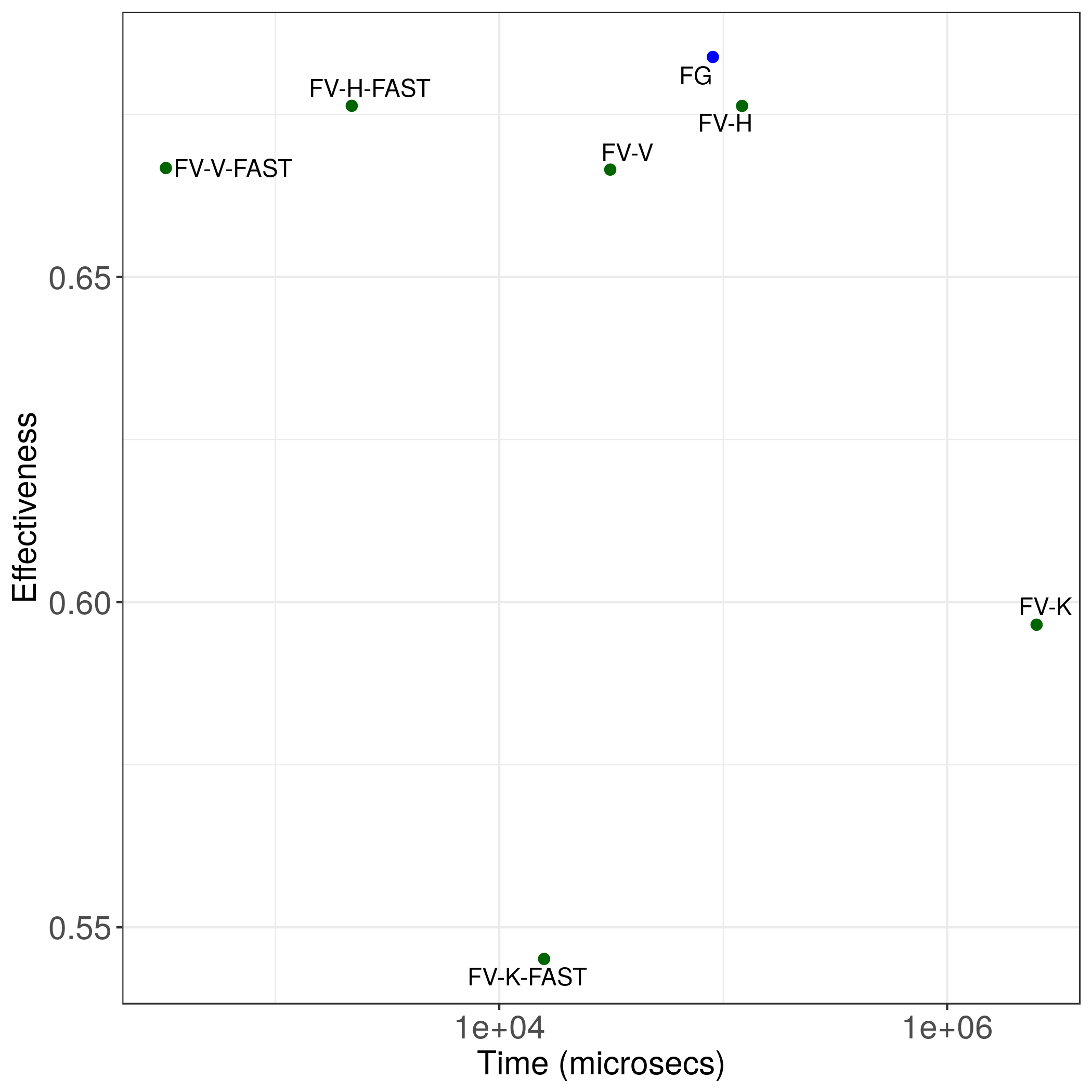}
\end{subfigure}
\begin{subfigure}{.33\textwidth}
\centering
\includegraphics[width=.97\linewidth]{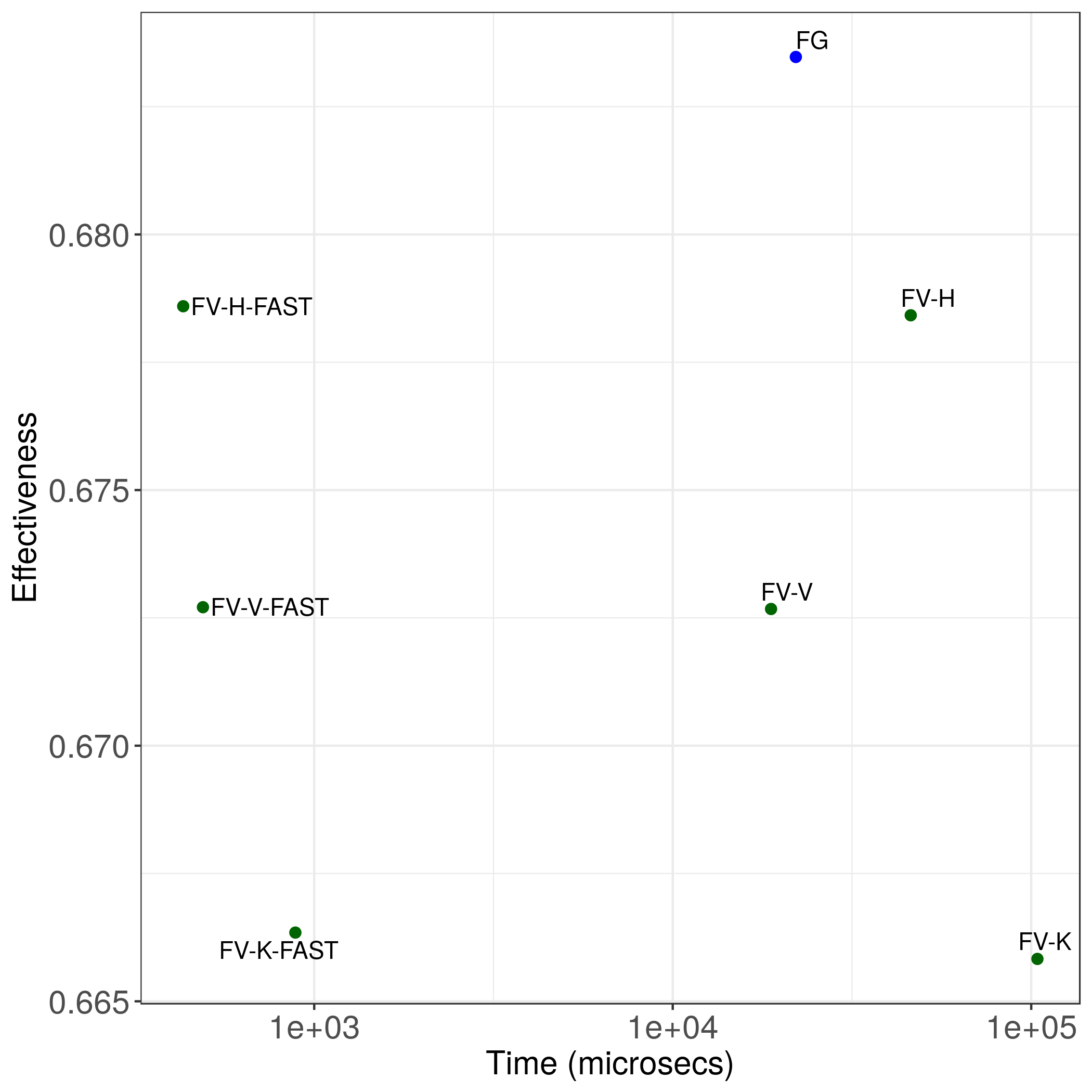}
\end{subfigure}
\begin{subfigure}{.33\textwidth}
\centering
\includegraphics[width=.97\linewidth]{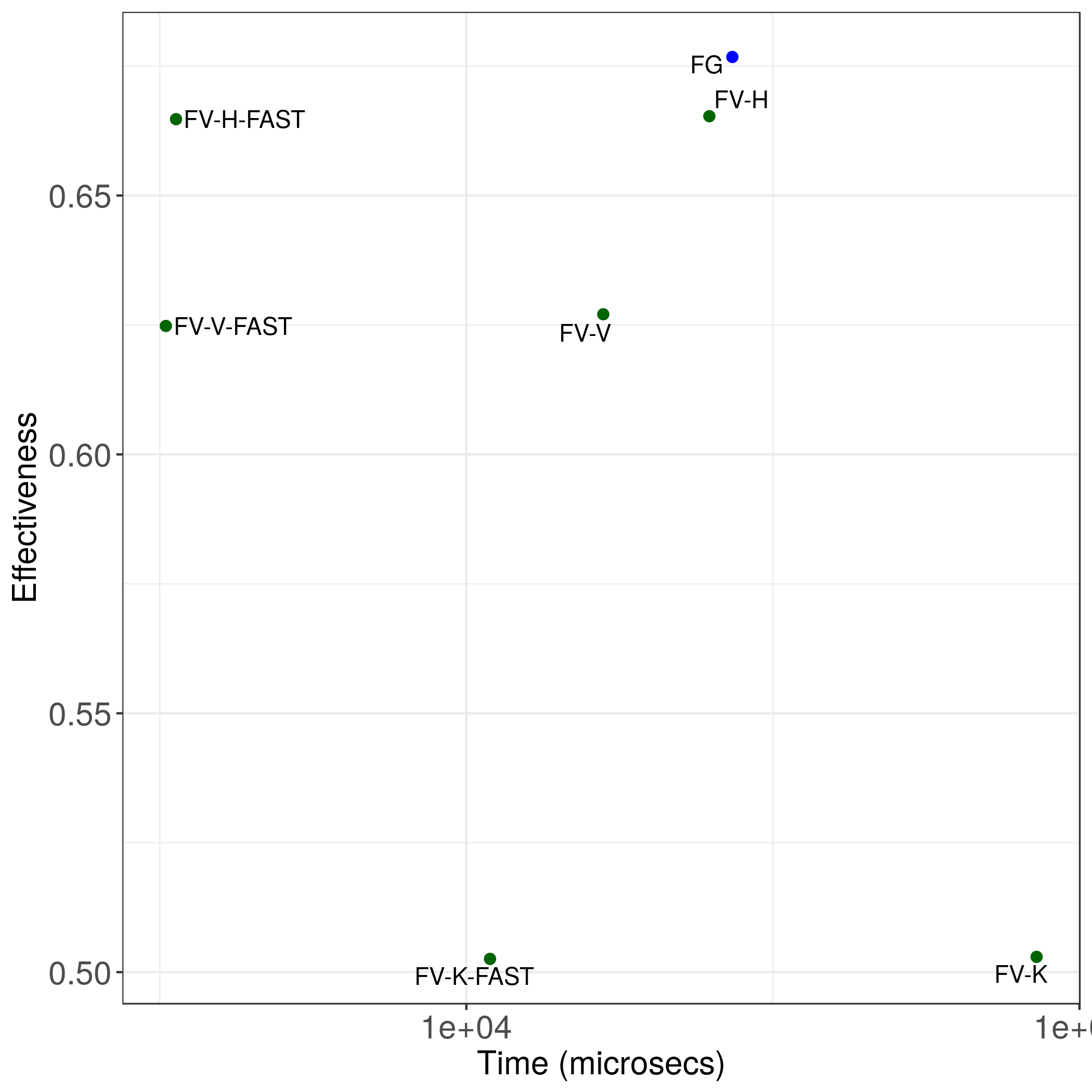}
\end{subfigure}
\caption{Effectiveness and efficiency trade-offs for FV and its embedding and indexed versions, in Ohsumed, for BoW-cosine + BoW-Jaccard + 2grams-cosine + 2grams-Jaccard + GNF-MCS + GNF-WGU + WMD, BoW-cosine + 2grams-cosine, and BoW-cosine + WMD, respectively.}
\label{fig:tradeoff-ohsumed}
\end{figure}

\begin{figure}[t!]
\centering
\begin{subfigure}{.33\textwidth}
\centering
\includegraphics[width=.97\linewidth]{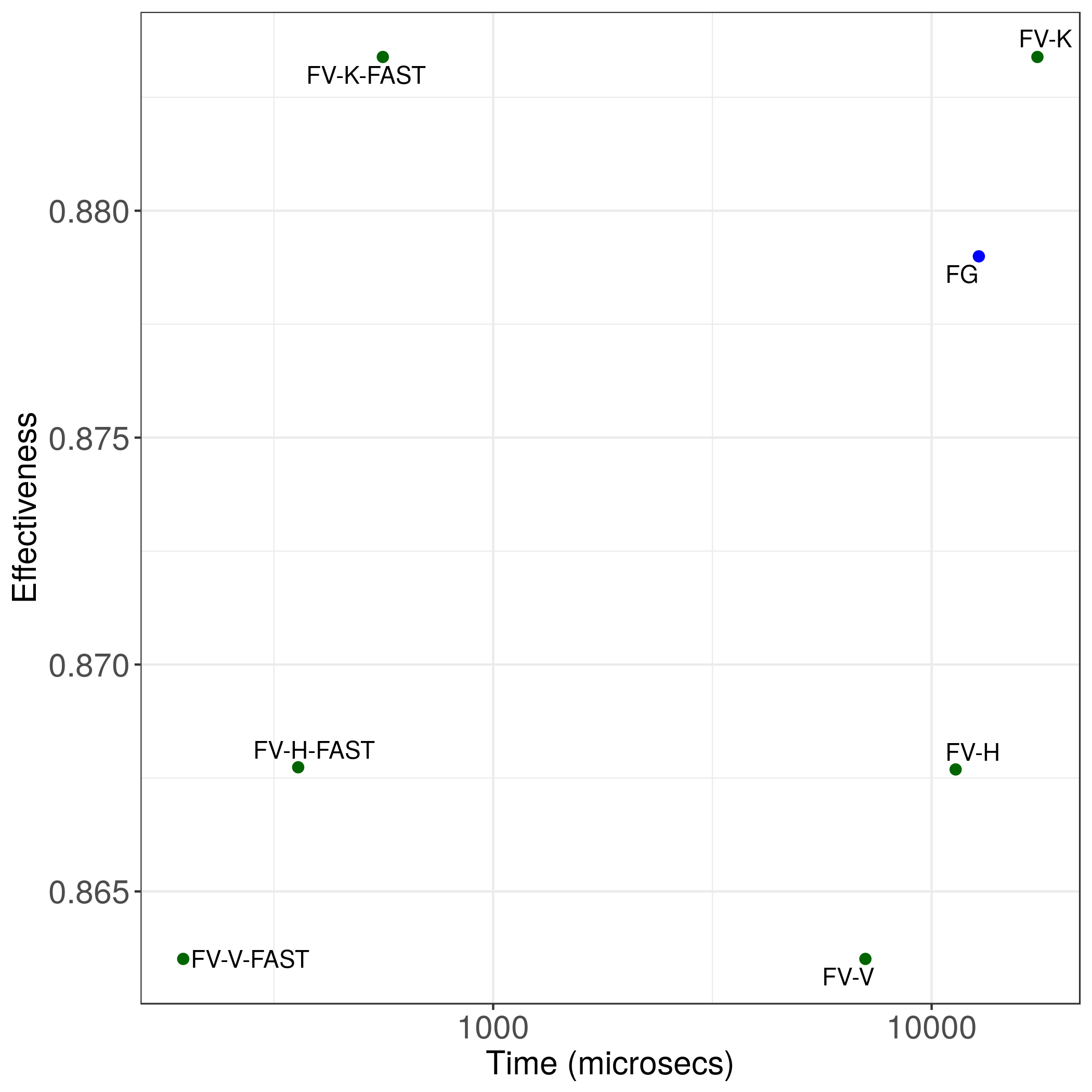}
\end{subfigure}
\begin{subfigure}{.33\textwidth}
\centering
\includegraphics[width=.97\linewidth]{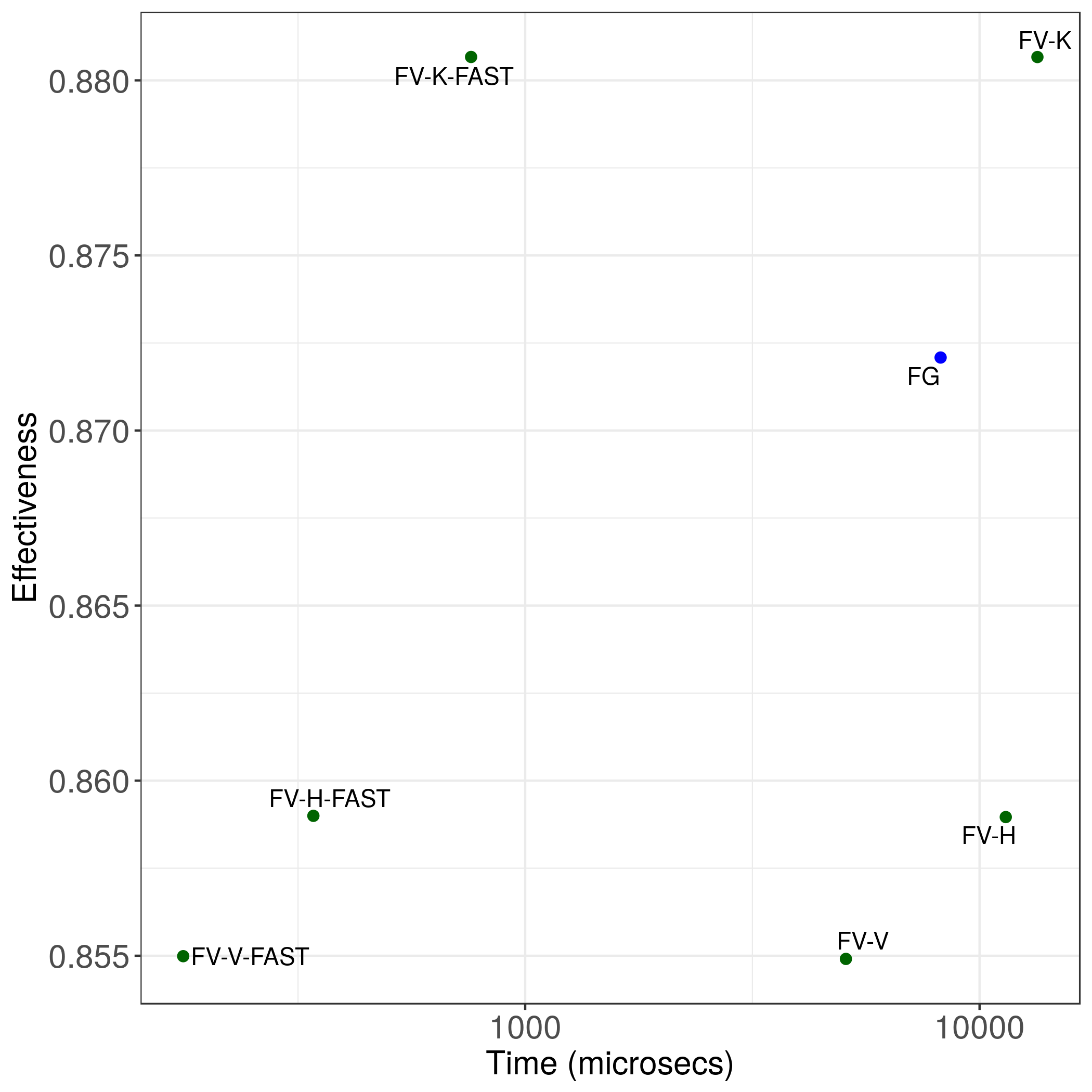}
\end{subfigure}
\begin{subfigure}{.33\textwidth}
\centering
\includegraphics[width=.97\linewidth]{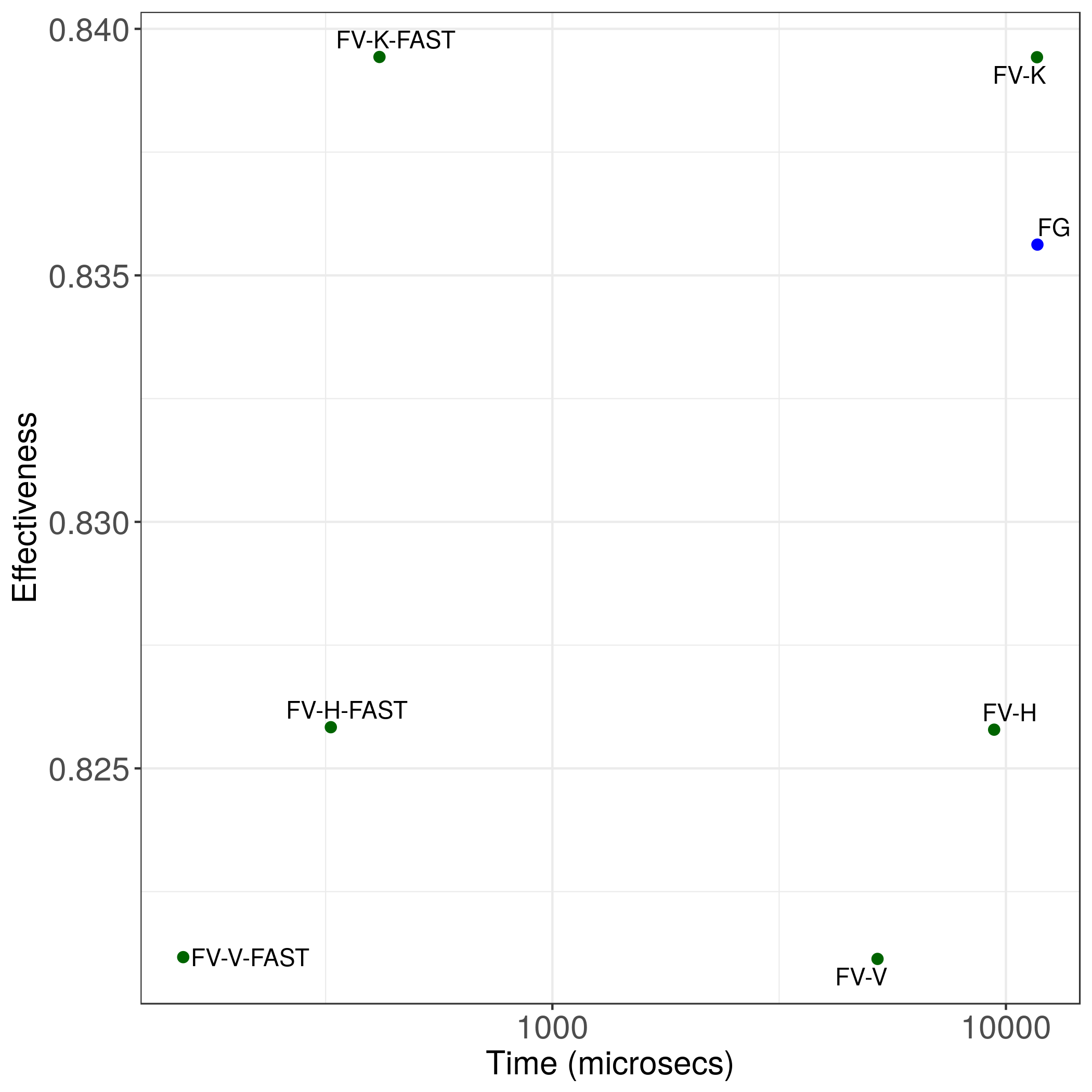}
\end{subfigure}
\caption{Effectiveness and efficiency trade-offs for FV and its embedding and indexed versions, in Brodatz, for LAS + CCOM + LBP, LAS + CCOM, and LAS + LBP, respectively.}
\label{fig:tradeoff-brodatz}
\end{figure}

\begin{figure}[t!]
\centering
\begin{subfigure}{.33\textwidth}
\centering
\includegraphics[width=.97\linewidth]{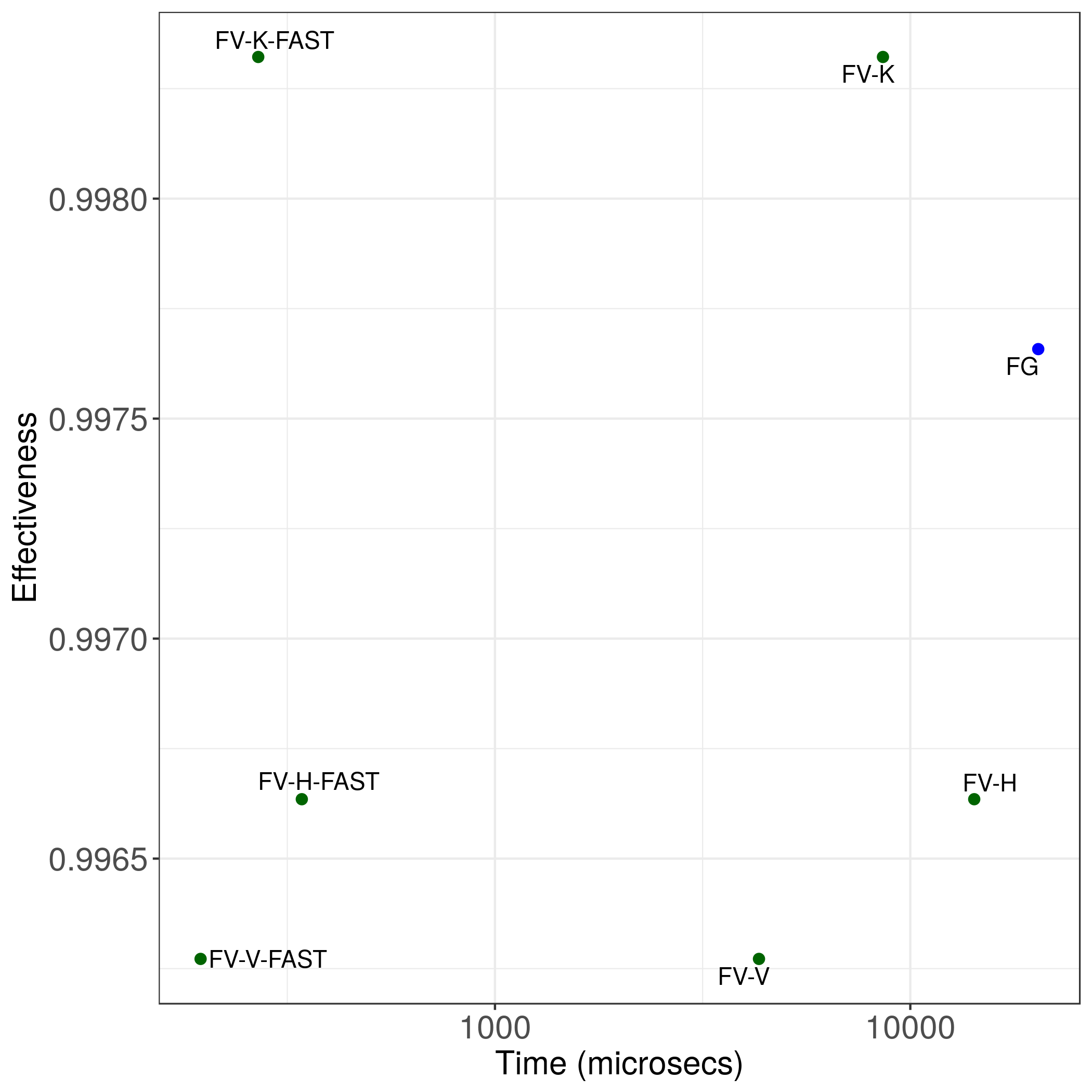}
\end{subfigure}
\begin{subfigure}{.33\textwidth}
\centering
\includegraphics[width=.97\linewidth]{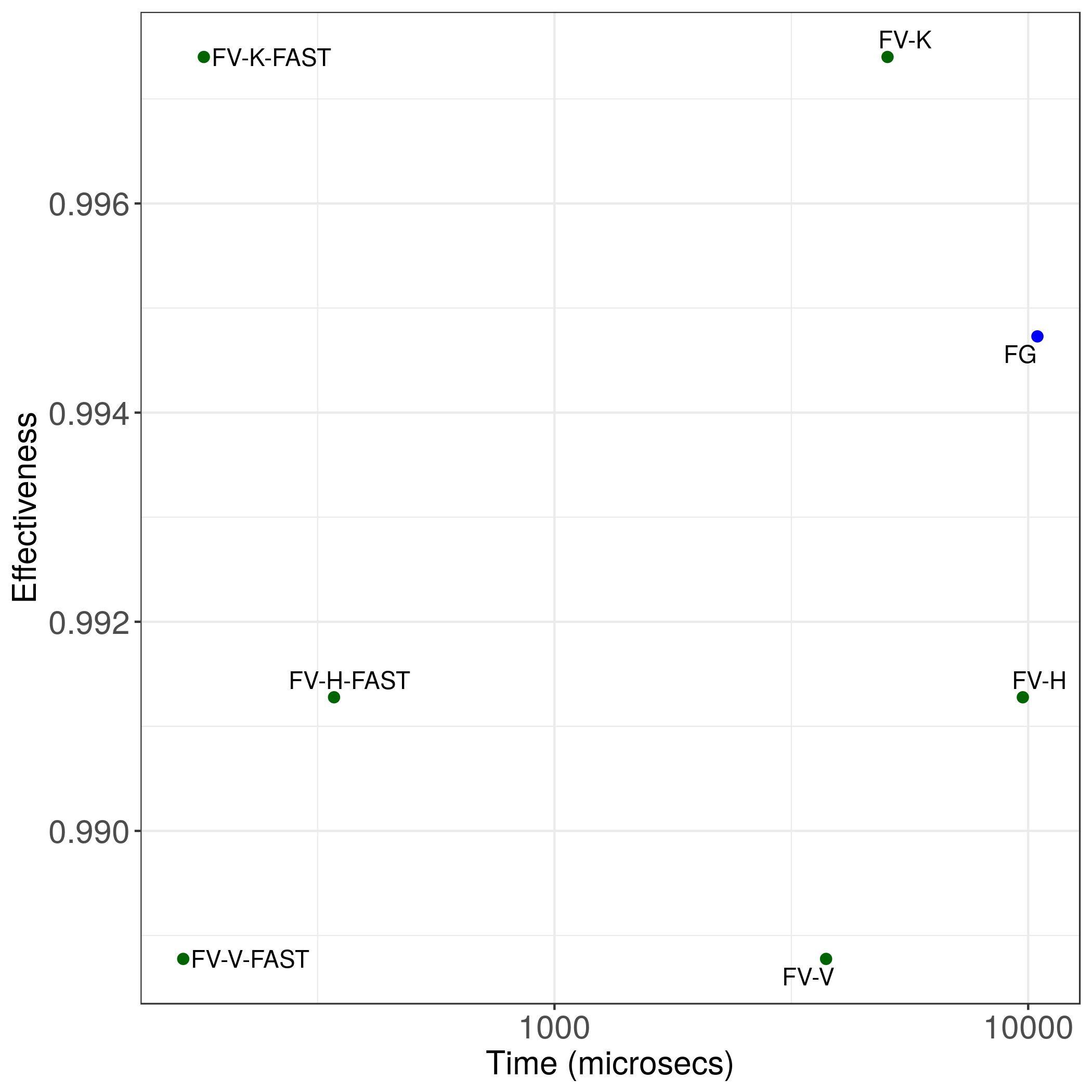}
\end{subfigure}
\begin{subfigure}{.33\textwidth}
\centering
\includegraphics[width=.97\linewidth]{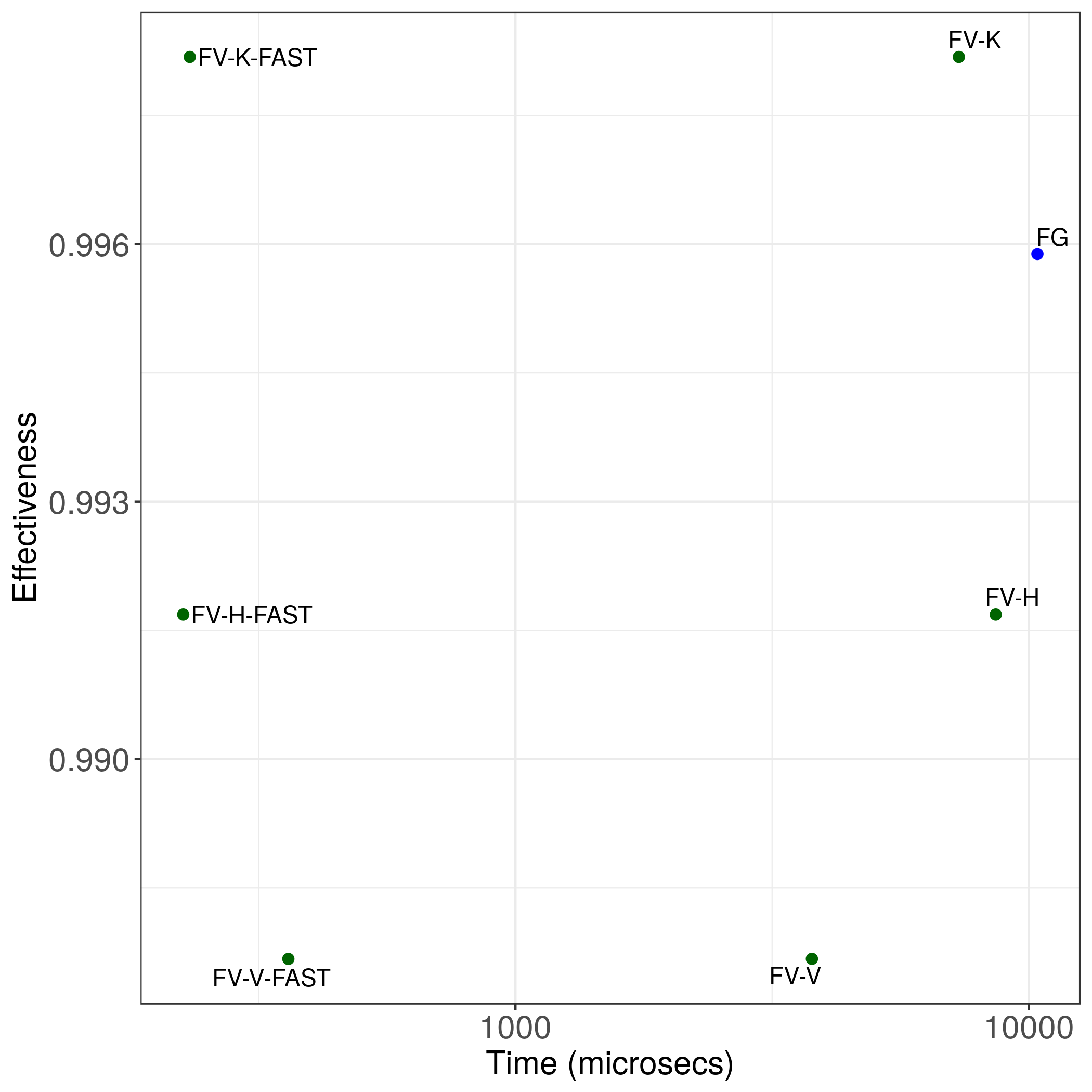}
\end{subfigure}
\caption{Effectiveness and efficiency trade-offs for FV and its embedding and indexed versions, in MPEG-7, for AIR + CFD + ASC + IDSC + BAS + SS, ASC + AIR, and AIR + CFD, respectively.}
\label{fig:tradeoff-mpeg7}
\end{figure}

\begin{figure}[t!]
\centering
\begin{subfigure}{.33\textwidth}
\centering
\includegraphics[width=.97\linewidth]{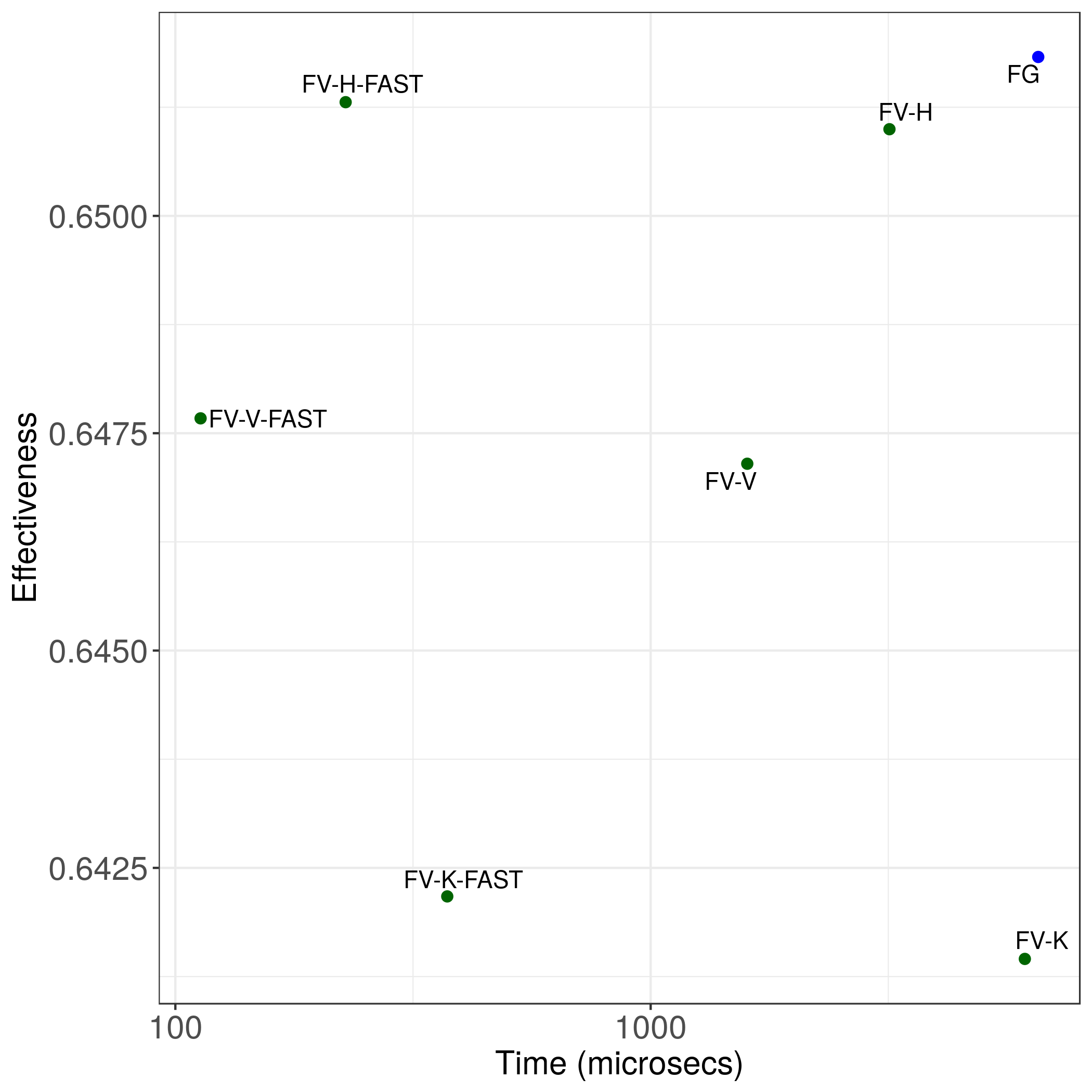}
\end{subfigure}
\begin{subfigure}{.33\textwidth}
\centering
\includegraphics[width=.97\linewidth]{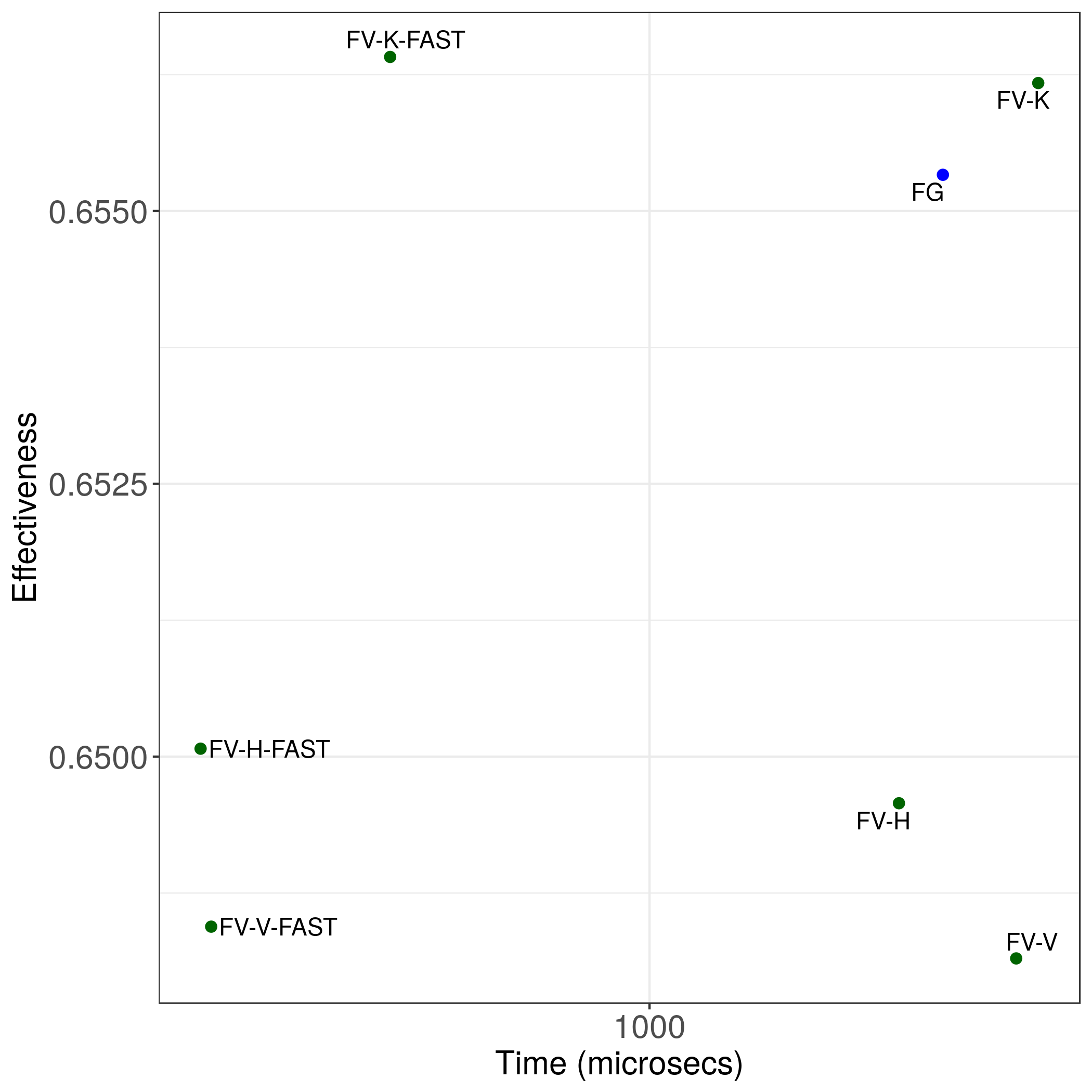}
\end{subfigure}
\begin{subfigure}{.33\textwidth}
\centering
\includegraphics[width=.97\linewidth]{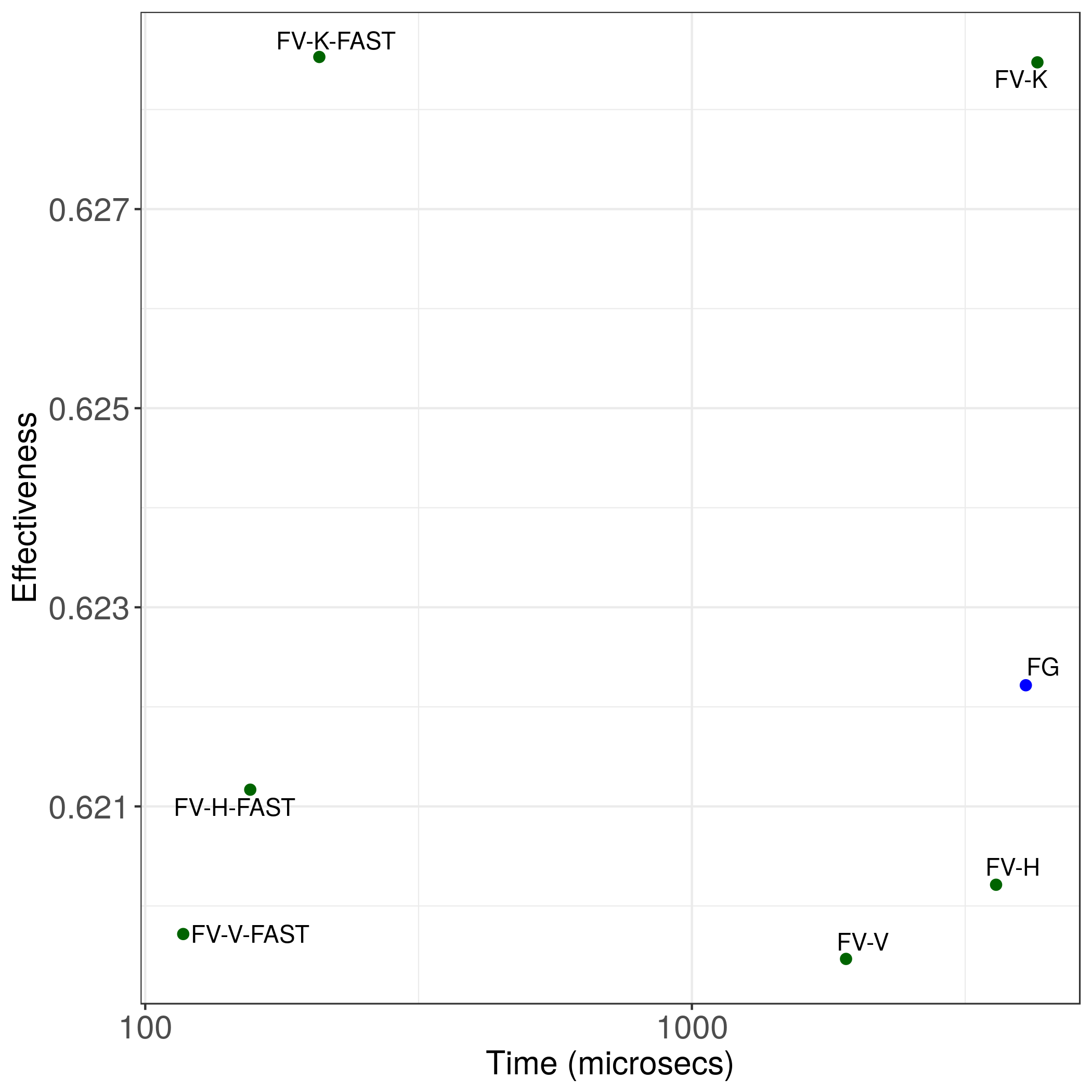}
\end{subfigure}
\caption{Effectiveness and efficiency trade-offs for FV and its embedding and indexed versions, in Soccer, for BIC + ACC + GCH, BIC + ACC, and BIC + GCH, respectively.}
\label{fig:tradeoff-soccer}
\end{figure}

\begin{figure}[t!]
\centering
\begin{subfigure}{.33\textwidth}
\centering
\includegraphics[width=.97\linewidth]{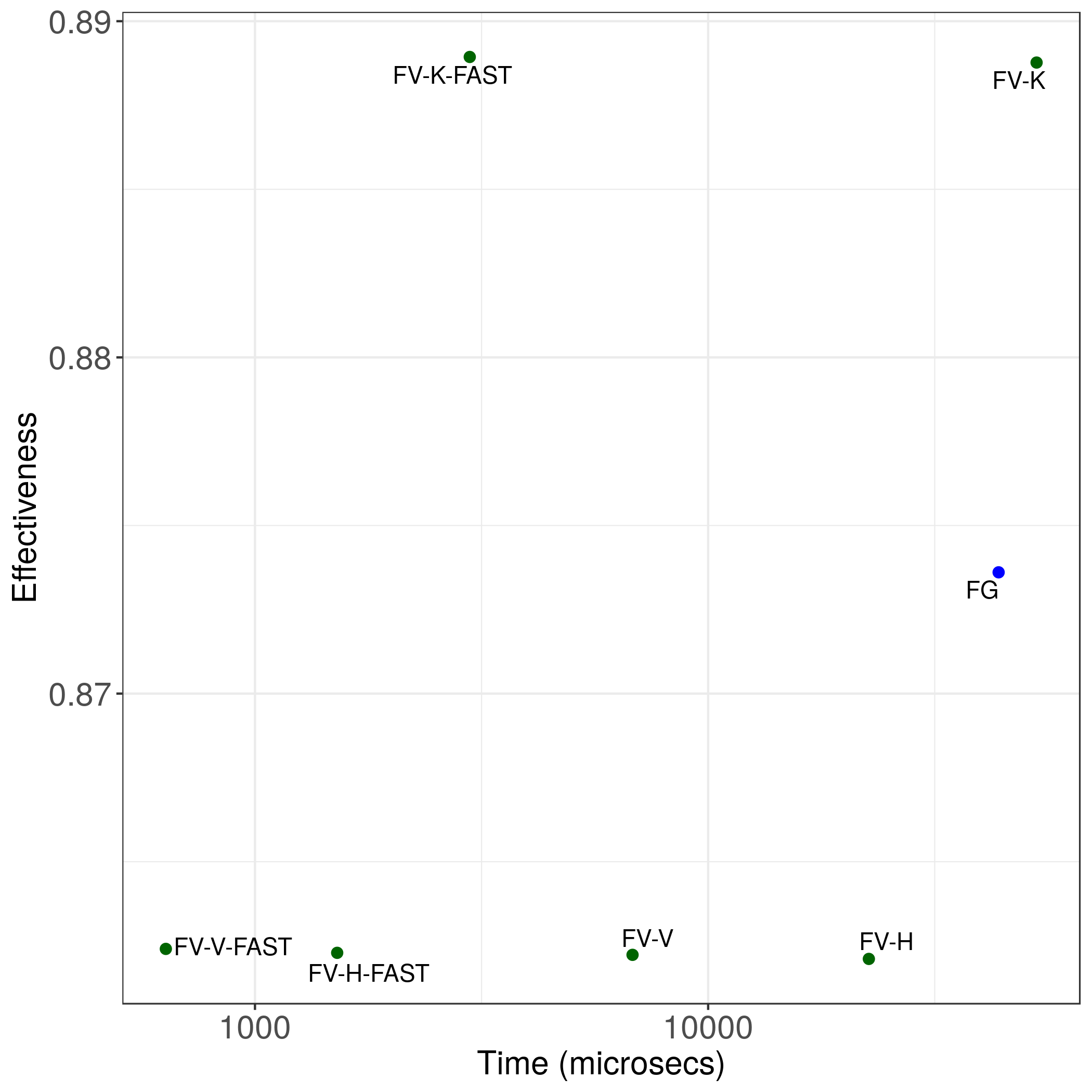}
\caption{}
\end{subfigure}
\begin{subfigure}{.33\textwidth}
\centering
\includegraphics[width=.97\linewidth]{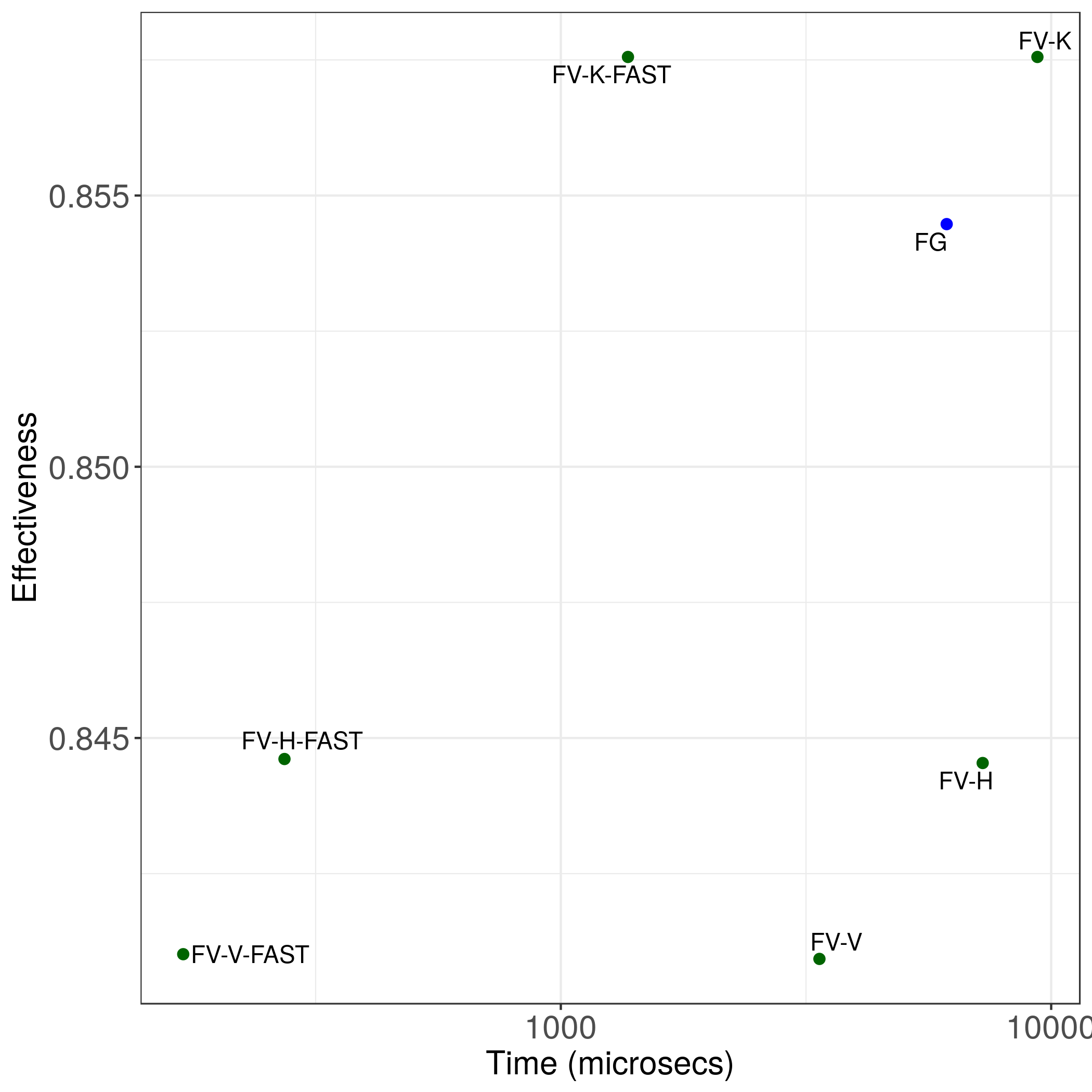}
\caption{}
\end{subfigure}
\begin{subfigure}{.33\textwidth}
\centering
\includegraphics[width=.97\linewidth]{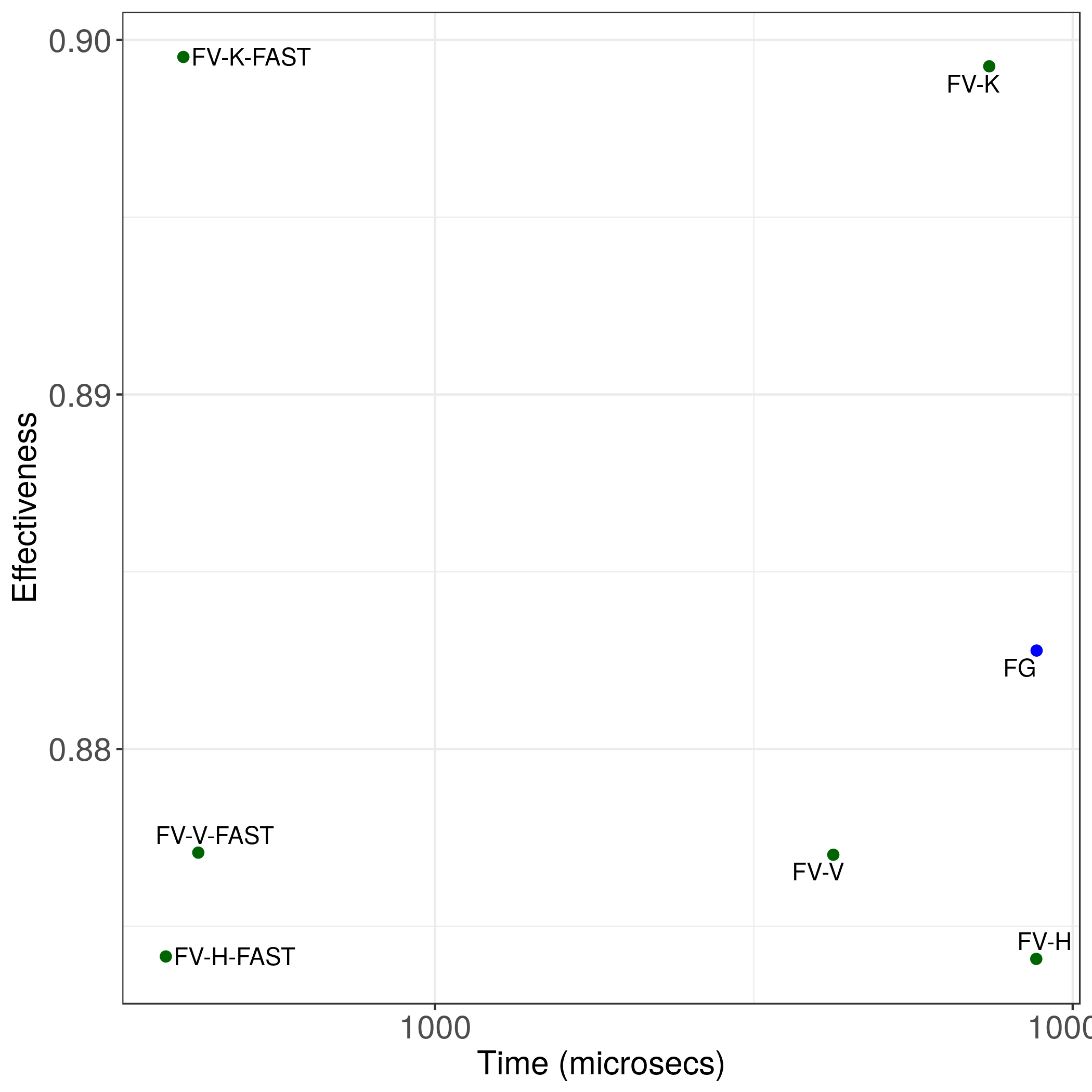}
\caption{}
\end{subfigure}
\caption{Effectiveness and efficiency trade-offs for FV and its embedding and indexed versions, in UW, for (a) JAC + BIC + DICE + BOW + OKAPI + JACCARD + TF-IDF + GCH + COSINE + LAS + HTD + QCCH, (b) JAC + BIC, and (c) JAC + OKAPI.}
\label{fig:tradeoff-uw}
\end{figure}

Figures~\ref{fig:tradeoff-ukbench},
\ref{fig:tradeoff-ohsumed},
\ref{fig:tradeoff-brodatz},
\ref{fig:tradeoff-mpeg7},
\ref{fig:tradeoff-soccer},
and \ref{fig:tradeoff-uw}
present the effectiveness scores related to the mean query times, for the three rank aggregations, in UKBench, Ohsumed, Brodatz, MPEG-7, Soccer, and UW, respectively. The times were measured on an Intel Core i7-7500U CPU @ 2.70GHz with 16GB of RAM.

\textit{FV} held much lower query times than \textit{FG} in all datasets, while preserving or surpassing its effectiveness in $4$ of $6$ datasets. Speedups from 10x to 100x were achieved by the indexed formulations. The gains are more significant for larger datasets, while also more demanded in those cases. Even for non-indexed FV approaches, the query times are considerably lower than \textit{FG}.

The efficiency for \textit{FV}, regardless of their indexed or non-indexed formulations, are affected by the number of dimensions, specifically the number of non-zero entries in the resulting fusion vectors as the dissimilarity functions can be designed for sparse vectors. In the non-indexed formulations, the collection size is the critical aspect for the final retrieval times, although also affected by the dimensionality.
In general, FV-V is the approach that conducts to the lowest dimensionality. FV-H presents high dimensionality, but actually not that high in terms of non-zero entries. FV-K, in practice, is the embedding that produced the fusion vectors with higher non-zero dimensions. That explains the lower efficiency in its retrieval when compared to the the other two approaches, although it is still much better than non-indexed \textit{FV} formulations and \textit{FG}, and also presents some effectiveness gains over its alternatives.

The experimental results achieved by \textit{FV} concerning effectiveness and efficiency shows a solid evidence that {\em fusion vectors} yield comparable or superior performance when compared with the state-of-the-art rank aggregation functions in effectiveness, while providing a fast alternative for aggregating lists in  search systems, a common shortcoming from previous works.

\section{Conclusions}
\label{secConc}

This paper introduced the concepts of embedding and indexing of graph-based rank aggregation representations, and their application for search systems.

Unsupervised embedding formulations were proposed and discussed, based on vertices, a hybrid of vertices and edges, and kernels.
The concept of fusion vectors was introduced, based on which a retrieval model could be established. The possibility of representing contextual information defined in terms of multiple ranked lists into a vector, opened the possibility of exploring indexing schemes. In this paper, we also investigated the use of approximate searches to deliver fast retrieval based on rank aggregation.

We demonstrated the flexibility of the proposed method in multimodal retrieval tasks, and evaluated the method experimentally across many diverse search scenarios, considering comparison with multiple state-of-the-art baselines.
Conducted experiments showed that our approach leads to comparable or superior results when compared with start-of-the-art rank aggregation functions considering effectiveness, while bringing a novel approach for fast retrieval on that context. The efficiency analysis showed a speedup improvement from $10$ to $100$ against our strongest baseline.
An extensive experimental section was conducted, considering $7$ recent related works, in $3$ distinct scenarios for each of the $6$ datasets investigated. Besides, evaluations conducted in UKBench dataset, against state-of-the-art aggregation methods, either based on early-fusion, late-fusion or supervised learning, showed that our unsupervised approach is not only effective and efficient but also very competitive to other approaches.

Future research directions concern the investigation of supervised or semi-supervised approaches for fusion graph embedding.
Another research direction aims at exploring incremental updating capabilities for the offline components of the solution, in order to reduce preprocessing costs.
Finally, we also plan to validate the concepts of fusion graphs and fusion vectors in other tasks, such as recommendation or classification.


\section*{References}

\bibliography{main}



\end{document}